\documentclass[12pt]{article}

\usepackage[T1]{fontenc}
\usepackage[total={6.5in,9in}]{geometry}
\usepackage{amsmath, amsthm, amssymb}
\usepackage{libertinus}
\usepackage{pifont}
\AtBeginDocument{\renewcommand{\checkmark}{\text{\ding{51}}}}
\usepackage[square,numbers,sort&compress]{natbib}
\usepackage{setspace}
\usepackage[bf]{caption}
\usepackage{titlesec}
\usepackage{booktabs}
\usepackage{graphicx}
\usepackage{xcolor}
\usepackage[export]{adjustbox}
\usepackage[hidelinks]{hyperref}
\usepackage{tabularx}
\usepackage{longtable}
\usepackage{fancyvrb}

\usepackage[nomarkers,nolists,noheads,tablesfirst]{endfloat}

\providecommand{\pipelineurl}{\url{https://github.com/forma-lab-mccombs/proforma-20q}}
\providecommand{\releaseurl}{\url{https://github.com/forma-lab-mccombs/forma-release}}

\providecommand{\apxref}[1]{Appendix~\ref{#1}}
\providecommand{\apxnoun}{appendix}

\providecommand{\apxstart}{\appendix}

\ifdefined\pdfpageattr    \let\rotatepagespecial\relax
\else\ifdefined\pdfvariable \let\rotatepagespecial\relax
\else \def\rotatepagespecial{}%
\fi\fi

\newlength{\wideexhibitwidth}
\providecommand{\wideexhibit}[1]{%
  \rotatepagespecial
  \setlength{\wideexhibitwidth}{0.93\textheight}%
  \rotatebox{90}{%
    \begin{minipage}{\wideexhibitwidth}%
      \setlength{\textwidth}{\wideexhibitwidth}%
      \setlength{\columnwidth}{\wideexhibitwidth}%
      #1%
    \end{minipage}}}

\providecommand{\benchname}{ProForma-20Q}
\providecommand{\Description}[1]{}

\titleformat{\section}[hang]{\normalfont\centering\bfseries}{\thesection.}{10pt}{}{}
\titleformat{\subsection}[hang]{\itshape}{\thesubsection.}{10pt}{}{}
\titlespacing{\subsection}{0pt}{*1}{1pt}
\titleformat{\subsubsection}[hang]{\itshape}{\thesubsubsection.}{10pt}{}{}
\titlespacing{\subsubsection}{0pt}{*1}{1pt}

\title{\Large\textbf{Long-Horizon Forecasting of Complete Financial Statements with \texorpdfstring{\textsc{Forma}}{Forma}}\\\vspace{0.25in}}
\author{\normalsize \MakeUppercase{Travis L. Johnson, Jiannan Jiang, Soumyabrata Chaudhuri,}\\
\normalsize \MakeUppercase{Yihao Chen$^{\dagger}$, Lauren Falvey$^{\dagger}$,} and \MakeUppercase{Donal O'Cofaigh$^{\dagger}$}\vspace{0.25in}}
\date{\normalsize \today}

\begin{document}

\maketitle

\begingroup
\renewcommand{\thefootnote}{}%
\footnotetext{All authors are with the University of Texas
at Austin, Austin, TX, USA. $^{\dagger}$These authors contributed equally to this research. We thank Quent Capital, the Langston Family Wealth Management Center, and the AIM Investment Center for financial support. Contact:
travis.johnson@mccombs.utexas.edu, jiannanjiang@utexas.edu,
schaudhuri@utexas.edu, yihaochen@utexas.edu, lmf2842@my.utexas.edu,
do7973@my.utexas.edu.}%
\endgroup

\begin{abstract}
Specialist training beats generalist scale when forecasting financial statements. To our knowledge, no prior work jointly forecasts complete financial statements beyond one year, yet in a discounted-cash-flow valuation most firm value sits past that window. We release \benchname{}, a reproducible benchmark for forecasting 78 statement line items 1--20 quarters ahead, for anonymized firms, from past statements and an industry code, scored by change-space $R^2$. On it, \textsc{Forma}, a transformer that reads statements as sets of (account, quarter, value) tuples and maximizes a masked-tuple Gaussian likelihood, beats every competitor we field: classical machine learning, chained gradient boosting, a zero-shot time-series foundation model, and frontier large language models. Its lead widens with horizon, where valuation needs accuracy most, and its Gaussian predictive intervals never under-cover. \textsc{Forma}'s forecasts nearly satisfy accounting identities; exact coherence is recoverable at no statistically significant accuracy cost. Its tuple interface supports scenario analysis without retraining, and we show that pinning future revenue paths sharpens the rest of the statement.


\end{abstract}



\clearpage
\pagenumbering{arabic}
\doublespacing

\section{Introduction}\label{sec:intro}

A complete \emph{pro forma} financial-statement forecast is the raw material of valuation, credit risk analysis, and financial planning. Yet the academic literature typically supplies only point forecasts of single items, usually earnings, a year ahead. In a discounted cash flow (DCF) valuation, by contrast, most of enterprise value sits in cash flows (not earnings) beyond a year. The largest component is terminal value: the firm's value at the forecast endpoint (typically 5--10 years), set as a peer or perpetuity multiple of a final-year forecast~\cite{wahlen2023financial}. A final-year number, however, is only as credible as the statement behind it: margins, reinvestment, and financing must hang together, which is why practitioners build complete pro formas. The economically relevant object is therefore a joint forecast of the whole statement at horizons where value lives. To our knowledge, no prior research provides such forecasts.

We introduce \benchname{}, a benchmark for this task. It asks models to predict 78 statement line items at horizons of 1--20 quarters using 12 quarters of lagged values for the same items and an industry code. We release the full protocol, which is reproducible with one command given WRDS access (\S\ref{sec:task}). Because valuation depends on aggregating conditional means across horizons, and we do not want to celebrate propagating persistent levels, models are ranked by out-of-sample $R^2$ for predicting changes.

Financial statement data resists the standard supervised-learning template. The average firm-quarter in our sample reports only 64 of the 78 line items, individual items are reported in as few as 32\% of firm-quarters, and \emph{fewer than 2\% of firm-quarters report the complete set}. The data also combines rigid structure (exact cross-account and cross-quarter accounting identities) with soft structure (latent economic states such as competitive moats and financial constraints).

We specify and evaluate \textsc{Forma}, a transformer-based architecture tailored to this setting. A statement history is encoded as a \emph{set} of (account, quarter, value) tuples, absent items contribute no token, and future items are masked tuples. \textsc{Forma} outputs a Gaussian predictive mean and variance for each mask and is trained to maximize the likelihood of the value behind the masks.

On \benchname{}, a $\sim$0.9M-parameter \textsc{Forma} beats all competitors (Table~\ref{tab:headline}): classical machine learning (ML) approaches using imputed missing values, such as penalized regressions, random forests (RF) \cite{breiman2001random,de2024noise}, and feed-forward neural networks (FFNNs) \cite{alberg2017fundamentals}; an in-suite re-implementation of its nearest conceptual competitor, the chained gradient boosted machine (GBM) of \citet{geertsema2026chained}; and two generalists, a zero-shot time-series foundation model (TSFM) and a best-effort panel of three frontier large language models (LLMs). \textsc{Forma} outperforms every model at every horizon beyond $h{=}2$ and its edge grows with forecast horizon, reaching at least 3.2 percentage points (pp) of $R^2$ by $h{=}20$ (Figure \ref{fig:horizon}). The LLMs' performance is generally poor: the best frontier model underperforms all but the simplest purpose-trained model, and its $R^2$ deficit relative to \textsc{Forma} grows from 5.5pp at $h{=}1$ to 15.9pp at $h{=}20$.

Our contributions are:

\begin{itemize}
\item \textbf{Task and protocol:} We define and release \benchname{}, a turnkey protocol for complete financial-statement forecasting, scored in change space on common samples.

\item \textbf{Architecture-to-problem fit:} We design a tuple-set transformer, \textsc{Forma}, around the data and task. Unreported items contribute no token (nothing is imputed); identity-aware masking forces the model to learn economics instead of accounting algebra; pinned-future masking trains it to condition on chosen realizations for scenario analysis.

\item \textbf{Probabilistic statement forecasts:} We produce predictive densities over the statement via a heteroskedastic head and five-seed mixture. These admit exact ex-post reconciliation at statistically insignificant accuracy cost, and Gaussian central intervals never under-cover at any horizon.

\item \textbf{The specialist-wins-and-widens result:} We document that \textsc{Forma} outperforms all competitors, including LLMs, and that its advantage widens with horizon. 
\end{itemize}
\section{Related Work}\label{sec:related}
\begin{table*}[t]
\wideexhibit{%
\caption{Academic research on forecasting firm-level financial statements. No surveyed system jointly forecasts the statement beyond one year, let alone distributionally; \textsc{Forma} does both.}
\label{tab:forecasting_comparison}
\centering
\scriptsize
\setlength{\tabcolsep}{3pt}
\renewcommand{\arraystretch}{1.15}
\begin{tabularx}{\textwidth}{@{}>{\raggedright\arraybackslash}p{3.70cm}>{\raggedright\arraybackslash}X>{\centering\arraybackslash}p{0.95cm}>{\centering\arraybackslash}p{1.00cm}>{\centering\arraybackslash}p{1.60cm}>{\centering\arraybackslash}p{1.45cm}>{\centering\arraybackslash}p{1.90cm}>{\centering\arraybackslash}p{1.90cm}@{}}
\toprule
Research type (representative papers) & Forecast outcome &
Outputs & \shortstack[c]{Horizon\\(Years)} & Architecture & \shortstack[c]{Cross-item\\modeling} &
\shortstack[c]{Distributional\\output} &
\shortstack[c]{Accounting\\structure} \\ 
\midrule
\multicolumn{8}{@{}l}{\textit{Panel A. Single-item earnings forecasts}} \\
\addlinespace[1pt]
Accounting-based earnings regressions
(\citet{hou2012implied,so2013predicting}) &
Annual earnings & 1 & 1--5$^{a}$ & OLS & --- & --- & --- \\
ML earnings models
(\citet{chen2022predicting,hess2023interpretable,campbell_expectations}) &
Earnings or earnings direction & 1 each & 1--5$^{a}$ & Trees, regressions, FFNNs & --- & --- & --- \\
Quantile earnings-risk model
(\citet{konstantinidi2016forecasting}) &
Distribution of annual earnings & 1 & 1 & Quantile regression & --- & $\checkmark$ & --- \\
\midrule
\multicolumn{8}{@{}l}{\textit{Panel B. Multi-item financial-statement forecasts}} \\
\addlinespace[1pt]
Multitask neural fundamentals
(\citet{alberg2017fundamentals,chauhan2020uncertainty}) &
Selected statement items & 16--17 & 1 & FFNN, LSTM & Shared model & \cite{chauhan2020uncertainty} & --- \\
Multi-model fundamentals benchmark
(\citet{divo2025forecasting}) &
Five selected items across all three statements &
1 or 5&
1 &
24 models$^{d}$ &
Varies &
Varies &
--- \\
Chained statements
(\citet{geertsema2026chained}) &
Partial income statement and balance sheet & $29 + 19^{b}$ & 1 & Chained GBMs & Sequential chain & --- & Articulated subset \\
\midrule
\multicolumn{8}{@{}l}{\textit{Panel C. Long-horizon complete statement forecasts (this paper)}} \\
\addlinespace[1pt]
\textbf{\textsc{Forma}} & \textbf{Standardized income statement, balance sheet, and cash-flow statement} &
\textbf{78} & \textbf{5} & \textbf{Tuple-set transformer} & \textbf{Shared model} & \pmb{$\checkmark$} &
\textbf{Optional projection$^{c}$} \\
\bottomrule
\end{tabularx}

\vspace{3pt}
\begin{minipage}{\textwidth}
\footnotesize
\textit{Notes:} \textit{Outputs} counts forecast quantities per system. \textit{Cross-item modeling} requires multiple output items modeled jointly or with shared parameters; statement variables used only as predictors do not qualify. A check denotes a continuous distributional forecast (binary class probabilities do not qualify); a dash denotes absent; ``Varies'' means the cited benchmark spans models with and without the property. $^{a}$Range across the cited studies. $^{b}$29 forecast directly; 19 computed from accounting relations. $^{c}$Identities can be imposed exactly ex post (\S\ref{sec:coherence}). $^{d}$From local statistical baselines to deep sequence models, including generic transformers (Transformer, TFT) and the pretrained Chronos.
\end{minipage}
}
\end{table*}


Table~\ref{tab:forecasting_comparison} and the paragraphs below contrast our work with existing research on firm-level financial forecasting across forecast scope, horizon, cross-item modeling, distributional output, and accounting structure. Single-outcome models reach multi-year horizons, whereas multi-output systems cover selected or stylized subsets for at most one year. To our knowledge, \textsc{Forma} is the first learned system to produce forecasts of a three-statement schema at valuation horizons, and the first to use a tuple-set transformer for this problem.\footnote{\citet{divo2025forecasting} include generic sequence transformers (Transformer, TFT, Chronos) consuming a dense 20-indicator panel to forecast five target items; none attends over sparse tuples, the representation that makes a full 78-item schema tractable.}

We also discuss three adjacent literatures---LLM forecasting, long-horizon time series, and coherent forecasting---that supply competing models and evaluation methods.

\paragraph{Earnings forecasting.} The accounting, finance, and ML literature typically forecasts \emph{earnings} or other individual items. Cross-sectional models in the \citet{hou2012implied} line~\cite{so2013predicting, li2014evaluating} and their ML successors~\cite{chen2022predicting, hess2023interpretable, jones2023machine} predict one target (a point realization or its direction of change) per fitted specification at one- to five-year horizons. Their performance relative to analyst consensus is specification-sensitive \cite{campbell_expectations}. Quantile regressions provide one-year-ahead distributional earnings forecasts \cite{konstantinidi2016forecasting}. Analyst forecasts and textbook pro formas are practice benchmarks, not learned systems~\cite{wahlen2023financial,park2025analysts}.

\paragraph{Statement-level systems.} \citet{alberg2017fundamentals} jointly forecast 16 fundamentals one year ahead with multi-task feed-forward and long short-term memory (LSTM) networks; \citet{chauhan2020uncertainty} extend the design to 17 targets with uncertainty estimates.  \citet{divo2025forecasting} benchmark 24 diverse models on five items spanning all three statements over four quarters. The closest system, \citet{geertsema2026chained}, forecasts 29 core income-statement and balance-sheet items one year ahead by chaining per-item GBMs and derives 19 more using accounting relations. We re-implement it in-suite and find it underperforms \textsc{Forma} on its own item footprint. 

\paragraph{LLMs and financial statements.} Large language models have been proposed as zero-shot forecasters that convert numeric series to token strings and extrapolate them directly~\cite{gruver2023llmtime}, though whether the language model itself helps numeric forecasting is disputed~\cite{tan2024language}. Our LLM panel (\S\ref{sec:results}) gives reasoning-enabled frontier models the same anonymized statements and industry code as every other model, a \emph{stronger} generalist condition than either the token-string prompting of \citet{gruver2023llmtime} or the fine-tuned adapters ablated by~\citet{tan2024language}. We nevertheless show that a small specialist dominates LLMs at every horizon in out-of-sample $R^2$.

\paragraph{Long-horizon and foundation-model forecasting.} 
The long-term time-series forecasting (``LTSF'') literature counts ``long term'' in steps, e.g. hundreds of hourly electric or traffic readings in \citet{zeng2023transformers}. We count ``long term'' in calendar time---our 20 quarterly steps span five years, far longer than standard LTSF horizons---so the difficulty is signal decay and distribution shift, not sequence length. Zero-shot time-series foundation models~\cite{das2024timesfm,ansari2025chronos2} supply our second generalist arm (\S\ref{sec:comparators}). Our classical baselines carry \citet{grinsztajn2022tree}'s trees-versus-deep-learning prior on tabular data: tree ensembles are the strongest competitors here too.

\paragraph{Coherent forecasting.} Forecast reconciliation adjusts predictions to satisfy aggregation constraints among hierarchical time series, as in the MinT method of \citet{wickramasuriya2019mint}, with end-to-end and probabilistic extensions~\cite{rangapuram2021end,panagiotelis2023}. As signed linear relations, accounting identities fall within the general framework of \citet{girolimetto2024}. We report unconstrained violations and apply variance-weighted reconciliation in \S\ref{sec:coherence}.

\section{\benchname{} Task and Protocol}\label{sec:task}
\subsection{Task}

Each forecasting example is indexed by a firm $f$ and a forecast origin quarter $t$; horizon $h$ refers to absolute quarter $t+h$. Let $\mathcal{D}$ denote the 78 accounting-item identifiers and $c_f$ the industry category of firm $f$. Each value the firm reports is a tuple $(h,\mathrm{id},x)$, where $\mathrm{id}\in\mathcal{D}$ identifies the accounting item and $x$ is its value at quarter $t+h$ after the standardization of \S\ref{subsec:data}.

The historical input is a set
\begin{align*}
S_{f,t} = \left\{ (h,\mathrm{id},x): h=-11,\ldots,0,\; \mathrm{id}\in\mathcal{D},\; \mathrm{id}\ \text{reported at } t{+}h \right\}.
\end{align*}
For any requested future horizon and accounting item, the forecasting task is to learn
\begin{align}
\widehat{x} = g(S_{f,t},c_f;h,\mathrm{id}), \qquad h=1,\ldots,20,\quad \mathrm{id}\in\mathcal{D},
\end{align}
where $\widehat{x}$ is the forecast of the corresponding standardized value at quarter $t+h$ given the information up to $t$.


\benchname{} thus fixes 12 quarters of history, 78 items, and a 20-quarter horizon. The formulation itself accommodates other item sets, histories, and horizons.

The tuple set is the task's canonical encoding: it records exactly what the firm reported and lets the reported subset vary freely across firm-quarters. Models need not consume it; the released protocol builds both the canonical \emph{tuple view} and a conventional \emph{tabular view} of the same panel for models  requiring rectangular inputs (\S\ref{sec:comparators}).\footnote{The tabular view's features include per-item recent levels and year-over-year changes on a fixed grid, with industry dummies and imputed missing values.} Forecasts from either view are scored identically.

\subsection{Data, Standardization, and Splits} \label{subsec:data}

The sample is quarterly U.S. filings (Compustat), excluding financial firms (SIC 6000--6999), whose statements comprise different line items with different economic meaning. Values are deflated by scale (origin-quarter $|\text{total liabilities}| + |\text{shareholders' equity}|$, which nearly always equals total assets), asinh-transformed, and standardized per (item, quarter) with statistics estimated from data available at the origin; standardized inputs are clamped at $|x|\le 6$.

Splits are temporal --- train 1971--2001, validation 2002--2009, test 2010--2024 --- with targets purged at each boundary: an example retains only target quarters realized within its own split, so nothing fit or selected before the test sample observes any post-2009 outcome. Every model conditions on the same 12-quarter reported history, origin-quarter scale, and Fama--French 48 industry (FF48)~\cite{fama1997industry}. None sees firm identity. We exclude stock returns, analyst forecasts, and other listed-firm features so that \textsc{Forma} can be applied to private firms. The panel spans 32{,}851 firms and 1{,}173{,}598 firm-quarters (609{,}269 train, 211{,}367 validation, 352{,}962 test), averaging 758 historical and 1{,}124 target tuples each.

The underlying data is licensed and cannot be redistributed, so we instead release a repository containing the full pipeline, configurations, evaluation code, and documentation of the complete item list, identities, standardization, and filtering procedures.\footnote{See \pipelineurl.} With it, anyone having WRDS access can rebuild the exact data environment in one command, verified by published checksums.

\subsection{Evaluation Using $R^2$ for Changes}\label{sec:eval}
We evaluate forecasts of \emph{changes} in standardized values: the truth is $x_{f, t}(h, \mathrm{id})-x_{f, t}(0, \mathrm{id})$ and a model's prediction is $\hat x_{f, t}^m(h, \mathrm{id})-x_{f, t}(0, \mathrm{id})$. Because the anchor $x_{f, t}(0, \mathrm{id})$ cancels out, mean absolute error (MAE) and mean squared error (MSE) are numerically identical in levels and changes; only the $R^2$ metric differs. Every $R^2$ in this paper reads as skill over assuming sample-average changes; in change space, $R^2$ of 20--40\% at multi-year horizons is strong, and level-space intuitions of 90\% or higher do not apply.

A DCF valuation sums discounted \emph{expected} cash flows across horizons. Medians do not add ($\mathrm{med}(A{+}B)\neq\mathrm{med}(A){+}\mathrm{med}(B)$), so median-targeted forecasts cannot feed a valuation. Our primary metric is therefore squared error in standardized space (Panel~A of Table~\ref{tab:headline}), which scores conditional-mean performance on the scale where items and firms are comparable. Because the asinh transform is nonlinear, recovering expected dollar cash flows requires the full predictive distribution.\footnote{Models with Gaussian predictive distributions, such as \textsc{Forma}, admit a closed form: when $X\sim N(\mu,\sigma^2)$ in standardized space, $\mathbb{E}[\sinh(aX+b)] = e^{a^2\sigma^2/2}\sinh(a\mu+b)$.} Distributional output is thus the bridge from standardized-space skill to dollar-space valuation, and we evaluate it when available (Panel~C of Table~\ref{tab:headline}).

We additionally report an absolute-error track (Panel~B of Table~\ref{tab:headline}), which scores conditional medians, for fairness to MAE-native models and to diagnose whether a model's poor $R^2$ reflects poorly located forecasts or implicit median-targeting.

\paragraph{Common samples.} Every comparison is evaluated on a common sample: a cell (firm, origin, item, horizon) contributes only if ground truth is available and every compared model predicts it. Models restrict coverage to different subsets of firms, items, origins, and/or horizons. A single all-model intersection would compound every such restriction and yield a severely limited sample. The protocol instead groups models by restriction and reports one footprint per group, re-scoring the full model suite inside each. Metrics are comparable within a footprint and never across footprints. Table~\ref{tab:headline} reports three such footprints; \S\ref{sec:comparators} states which restriction binds each comparator. Finally, a cell at horizon $h$ is scoreable only if the firm reports at $t{+}h$, so long-horizon evaluation conditions on realized survivors; comparisons remain fair (identical cells for all models), but level interpretations carry that caveat.

\paragraph{Significance.} We measure statistical significance by \citet{diebold1995comparing} tests, accounting for dependence across cells in the same quarter by collapsing to calendar-quarter means (the effective sample is the 60 test quarters, not 327M cells), with \citet{newey1987simple} standard errors using bandwidth 19 for overlapping origins, and the \citet{harvey1997testing} (HLN) small-sample correction; positive statistics favor the comparator.

\section{\textsc{Forma}: A Tuple-Set Transformer}\label{sec:model}

\paragraph{Tuple-set representation.} Fix a firm-origin example $(f,t)$. Each observed historical tuple $(h,\mathrm{id},x)\in S_{f,t}$ is a token, and each requested future pair $(h,\mathrm{id})$ is a query token whose value is hidden from the encoder. The initial representation of a token is
\begin{align}
z_{h,\mathrm{id}}^{(0)} = E_{\mathrm{acct}}(\mathrm{id})+E_{\mathrm{horizon}}(h)+E_{\mathrm{value}}(x,m),
\end{align}
where $m$ indicates whether the token's value is hidden and
\begin{align}
E_{\mathrm{value}}(x,m)=
\begin{cases}
w_x x, & m=0,\\
e_{\mathrm{mask}}, & m=1.
\end{cases}
\end{align}
Here $E_{\mathrm{acct}}$ is a learned account embedding,
$E_{\mathrm{horizon}}$ a fixed sinusoidal encoding of the relative quarter $h$, and $w_x$ a learned projection for the standardized value.

An unreported historical item contributes no tuple and hence no token, while masked historical observations and query tokens remain in the set with value $e_{\mathrm{mask}}$. Missingness is therefore native to the representation: incomplete statements can be used without imputation, avoiding complete-case selection and the researcher degrees of freedom afforded by an imputation rule --- a choice every tabular model must make (\S\ref{sec:comparators}).

Two context tokens complete the input: the firm's industry, $z_{\mathrm{ind}}^{(0)}=E_{\mathrm{industry}}(c_f)+E_{\mathrm{horizon}}(0)$, and the origin-quarter scale deflator of \S\ref{subsec:data}, standardized like any other value (without self-deflation); neither is a prediction target. Tokens are stacked unordered to form $H^{(0)}$, and a Transformer encoder~\cite{vaswani2017attention} attends over the set as in \citet{lee2019set}, so every token conditions on the full context. The encoder has 4 layers, $d_{\mathrm{model}}=128$, and 4 attention heads ($\approx$0.9M parameters).

\paragraph{Training objective.}\label{sec:masking}
Training uses masked prediction over the tuple set. The loss is evaluated on masked accounting tokens with observed targets, and masking has two designed variations.

\emph{(a) Identity-aware grouped masking.}
We randomly mask some reported historical tuples for regularization. However, statement values are tied by exact linear identities, so masking single tuples could teach constraint algebra rather than economics. When masking touches a complete identity group we therefore mask at least two of its members; tuples with no complete identity instance in the tuple set are randomly masked as singletons.

\emph{(b) Pinned-future masking.}
Half the training examples mask the entire future (pure forecasting); the other half reveal ${\approx}5\%$ of reported future tuple values as inputs. The model thereby learns to condition on partial future information, the capability behind the scenario analysis of \S\ref{sec:scenario}. 

The future query grid is constructed ex ante and firm-uniform, so that slot existence never depends on which cells the firm later reports. Otherwise the mere pattern of slots would reveal account additions, removals, or exit from the sample---a survivorship and look-ahead leak whose closure cost ${\approx}0.7$pp of test $R^2$.

\paragraph{Direct multi-horizon probabilistic forecasts.} For each requested pair $(h,\mathrm{id})$, we map $u=\operatorname{Concat}\big[z_{h,\mathrm{id}}^{(L)}, E_{\mathrm{horizon}}(h)\big]$ to a location and a heteroskedastic scale:
\begin{align}
\mu_{f,t}(h,\mathrm{id})=f_{\mu}(u), \qquad \log \sigma_{f,t}^{2}(h,\mathrm{id}) = f_{\sigma}(u),
\end{align}
with $\mu_{f,t}(h,\mathrm{id})$ serving as the task's point forecast $\widehat{x}$. We implement each mapping as a two-layer multilayer perceptron (MLP). 

The primary head is a heteroskedastic Gaussian $(\mu,\sigma)$, trained with the $\beta$-NLL loss ($\beta=0.5$)~\cite{seitzer2022pitfalls}, under a horizon curriculum; for the MAE track we also train a Laplace head $(\mu,\, b=\sigma/\sqrt{2})$ with the corresponding $\beta$-NLL.

We train five seeds and treat their predictive distributions as an equal-weight mixture. Distributional metrics use the exact mixture rather than averaged parameters: NLL uses the mixture density, CRPS a closed-form expression for the mixture family.\footnote{For the Gaussian mixture, we use \citet{grimit2006crps}; for the Laplace mixture, we use the analogous expression obtained from $\mathrm{CRPS}=\mathbb{E}|X-y|-\tfrac{1}{2}\mathbb{E}|X-X'|$.}

\section{Results}\label{sec:results}

\begin{table*}[t]
\caption{\benchname{} test-set results by loss geometry. Panel~A scores squared error (change-space $R^2$) on conditional-mean forecasts, Panel~B absolute error (MAE) on median forecasts, Panel~C proper scores on predictive densities. Columns are exact common samples---Full, Geert.\ \cite{geertsema2026chained}, and LLM; compare within columns, never across. Dashes: no forecast on that sample (for Chronos-2's NLL, no usable log score; see text). Significance: Diebold--Mariano vs.\ \textsc{Forma} within column (quarter-clustered, Newey--West, HLN-corrected); $^{*}/^{**}/^{***}$ \textsc{Forma} better at 10/5/1\%, $^{\dagger}$ comparator better. Model specifications: \S\ref{sec:comparators}.}
\label{tab:headline}
\small
\begin{minipage}[t]{0.53\textwidth}\vspace{0pt}%
\centering
\begin{tabular}{lccc}
\multicolumn{4}{l}{\emph{Panel A: squared-error track --- change-space $R^2\!\uparrow$ (conditional-mean forecasts)}}\\
\toprule
Model & \shortstack[c]{Full sample\\327.2M} & \shortstack[c]{Geert.\ sample\\109.1M} & \shortstack[c]{LLM sample\\2.15M} \\
\midrule
\textbf{\textsc{Forma} (Gaussian, 5-seed)} & \textbf{0.289} & \textbf{0.247} & \textbf{0.299} \\
Random Forest            & 0.272$^{***}$ & 0.231$^{***}$ & 0.279$^{***}$ \\
Elastic Net              & 0.258$^{***}$ & 0.217$^{***}$ & 0.269$^{***}$ \\
FFNN (linear, 5-seed)    & 0.253$^{***}$ & 0.207$^{***}$ & 0.264$^{***}$ \\
FFNN (large, 5-seed)     & 0.247$^{***}$ & 0.200$^{***}$ & 0.253$^{***}$ \\
Fade / AR(1)             & 0.183$^{***}$ & 0.170$^{***}$ & 0.179$^{***}$ \\
Chronos-2 (mean)         & 0.155$^{***}$ & 0.108$^{***}$ & 0.152$^{***}$ \\
Seasonal random walk     & $-0.041^{***}$ & $-0.068^{***}$ & $-0.041^{***}$ \\
\midrule
Chained GBM (MSE)~\cite{geertsema2026chained} & --- & 0.185$^{***}$ & --- \\
Claude Opus 4.8          & --- & --- & 0.186$^{***}$ \\
GPT-5.5                  & --- & --- & 0.174$^{***}$ \\
Claude Sonnet 5          & --- & --- & 0.158$^{***}$ \\
\bottomrule
\end{tabular}
\end{minipage}\hfill
\begin{minipage}[t]{0.44\textwidth}\vspace{0pt}%
\centering
\setlength{\tabcolsep}{3.5pt}%
\begin{tabular}{lccc}
\multicolumn{4}{l}{\emph{Panel B: absolute error --- MAE$\downarrow$ (median)}}\\
\toprule
Model & Full & Geert. & LLM \\
\midrule
\textsc{Forma} (Laplace, 5-seed) & \textbf{0.369} & \textbf{0.364} & \textbf{0.348} \\
Chronos-2 (median)       & 0.431$^{***}$ & 0.418$^{***}$ & 0.407$^{***}$ \\
Seasonal random walk     & 0.445$^{***}$ & 0.426$^{***}$ & 0.425$^{***}$ \\
\midrule
Chained GBM (L1)~\cite{geertsema2026chained} & --- & 0.400$^{***}$ & --- \\
Claude Opus 4.8          & --- & --- & 0.362$^{**}$ \\
GPT-5.5                  & --- & --- & 0.363$^{**}$ \\
Claude Sonnet 5          & --- & --- & 0.368$^{***}$ \\
\bottomrule
\end{tabular}

\medskip
\begin{tabular}{lcccc}
\multicolumn{5}{l}{\emph{Panel C: density track (full sample)}}\\
\toprule
Model & NLL$\downarrow$ & CRPS$\downarrow$ & Cov$_{50}$ (0.50) & Cov$_{90}$ (0.90) \\
\midrule
\textbf{\textsc{Forma} (Laplace)} & \textbf{0.160} & \textbf{0.293} & 0.637 & \textbf{0.916} \\
\textsc{Forma} (Gaussian) & 0.660 & 0.318 & 0.726 & 0.937 \\
FFNN (large)             & 0.842 & 0.349 & 0.700 & 0.939 \\
FFNN (linear)            & 0.992 & 0.369 & 0.781 & 0.952 \\
Chronos-2                & --- & 0.336 & \textbf{0.509} & 0.875 \\
\bottomrule
\end{tabular}
\end{minipage}

\end{table*}

\subsection{Competing Models}\label{sec:comparators}

Every comparator receives the information set of \S\ref{sec:task} and is scored in its native track. The tabular baselines (elastic net, RF, and two FFNNs) consume the benchmark's tabular view: four recent levels and eight year-over-year changes per item plus industry dummies, with missing features imputed via the local XS estimator of \citet{bryzgalova2025missing}.\footnote{The imputed feature matrix itself contains lagged levels and changes, so backward-looking information enters the factor model in the spirit of their local B-XS variant.} Each forecasts every (item, horizon) pair directly rather than recursively, so comparisons isolate the pipeline, not forecast strategy; the penalized regression and RF fit a separate model per pair, while the FFNNs and \textsc{Forma} share parameters across all pairs through a vector output head. The FFNNs are five-seed ensembles with the same heteroskedastic head and mixture treatment as \textsc{Forma}, so they also enter the density track.

Two simple baselines anchor the table. The seasonal random walk repeats each item's most recent value from the same fiscal quarter, the standard naive expectation for quarterly accounting series~\cite{foster1977quarterly,bernard1990evidence}. The fade/AR(1) baseline fits one pre-test OLS regression per item and horizon, pooled across firms, on the current level, capturing mean reversion.

The chained GBM of \citet{geertsema2026chained} is the only prior model we are aware of that forecasts a sizable block of the statement. We re-implement it under our common protocol using the 12-quarter tabular feature view rather than the original paper's single annual lag. We estimate the original \(L_1\) specification for the absolute-error track and an \(L_2\) variant for the squared-error track. Because 25 of the chained items lie in the \benchname{} universe, comparisons use the corresponding item footprint (the ``Geert.\ sample'').

The generalists come in two flavors. We prompt three frontier LLMs (Claude Opus 4.8, GPT-5.5, Claude Sonnet 5; all with extended thinking enabled) with the firm's industry and the 12-quarter anonymized history as dollar values. They forecast dollar values of the statement primitives at $h{=}1$--$20$; we reconstruct the remaining items via the accounting identities, apply the standardization of \S\ref{sec:task}, and score on a calendar-balanced 2{,}103-origin subsample (the ``LLM sample''). Each model runs two prompt arms; Table~\ref{tab:headline} reports the better arm per model, and our full prompts are available in the release documentation.\footnote{See \releaseurl.} The LLMs emit point forecasts without a stated estimand, so we score them in both point tracks.

The zero-shot TSFM Chronos-2~\cite{ansari2025chronos2} receives the same 12-quarter context in raw dollar values (it applies its own normalization), treating the 78 items as one multivariate group with missing values handled natively. It emits 21 native quantiles, which we map into benchmark space through the origin-frozen transform. Each track scores Chronos-2 at its matched estimand: the quantiles are integrated (trapezoidal, flat tails) to a conditional mean for the squared-error track, read at the median for the absolute-error track, and scored as a predictive distribution in the density track.

\subsection{Forecast Performance}

\paragraph{Full sample.} Panel~A of Table~\ref{tab:headline} presents the headline: \textsc{Forma} explains 28.9\% of the cross-sectional variance of realized changes, ahead of the RF (27.2\%), penalized regression (25.8\%), and both FFNN baselines (25.3\% and 24.7\%). Every gap is significant at the 1\% level under quarter-clustered DM tests.

\begin{figure}
\centering
\includegraphics[width=\columnwidth]{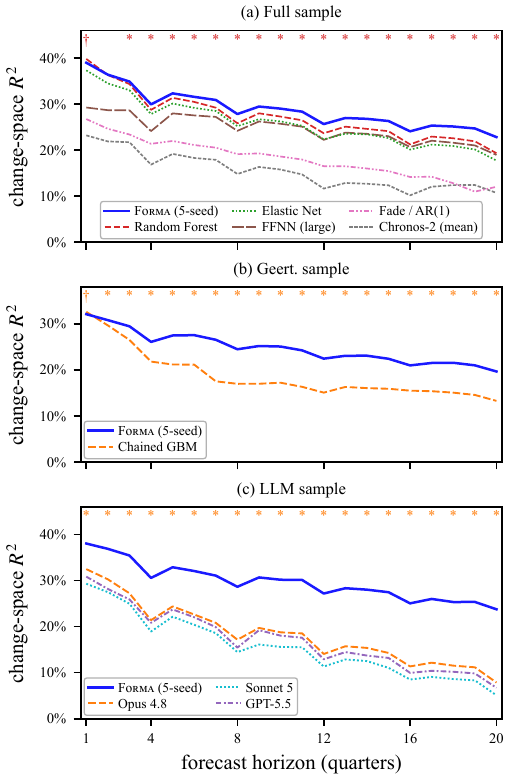}
\Description{Three stacked panels showing change-space R-squared by forecast horizon: (a) the full sample, \textsc{Forma} versus five baselines; (b) the Geert.\ common sample, \textsc{Forma} versus the chained GBM; (c) the LLM common sample, \textsc{Forma} versus three frontier LLMs.}
\caption{Change-space $R^2$ by forecast horizon on (a)~the full sample (\textsc{Forma} vs.\ five baselines), (b)~the Geert.\ common sample (\textsc{Forma} vs.\ the chained GBM), and (c)~the LLM common sample (\textsc{Forma} vs.\ three frontier LLMs). Markers atop each panel show quarter-clustered DM significance at the 5\% level against RF, the GBM, and Opus 4.8 (the strongest LLM), respectively ($\dagger$~comparator better, $*$~\textsc{Forma} better).}
\label{fig:horizon}
\end{figure}

Figure \ref{fig:horizon} shows that the RF slightly outperforms \textsc{Forma} one quarter ahead ($R^2$ 39.8\% vs.\ 39.0\%; DM $t{=}{+}12.3$), but the models are at parity at $h{=}2$, and from $h{=}3$ \textsc{Forma} is ahead with significance growing through $h{=}20$ (0.225 vs.\ 0.193; $t$ reaching $-18.6$). RF may therefore be better suited to short-term earnings timing, while \textsc{Forma} suits medium- and long-term tasks such as valuation or credit risk analysis.

The mechanism is visible in the baseline ladder. Fundamentals mean-revert at conditional, item-specific rates~\cite{nissim2001ratio}. The seasonal random walk misses reversion entirely ($R^2{=}{-}4.1\%$). The pooled fade\slash AR(1) baseline captures \emph{unconditional} reversion and explains 18.3\% of the variation, over half of \textsc{Forma}'s total. What separates the models is the \emph{conditional} component---reversion speeds that depend on the rest of the statement. That is also where capacity alone fails: the larger FFNN ($\approx$4.2M parameters) \emph{under}performs its linear sibling (24.7\% $R^{2}$ vs.\ 25.3\%; $\approx$3.1M parameters), and both trail the $\approx$0.9M-parameter transformer by 3.6--4.2 percentage points. Architecture, not capacity, drives \textsc{Forma}'s edge.

\paragraph{Chained GBM} In Panel A, the \(L_2\) GBM variant achieves an \(R^2\) of 18.5\%, compared with 24.7\% for \textsc{Forma} on their common sample. On the absolute-error track, the original \(L_1\) specification of \citet{geertsema2026chained} posts an MAE of 0.400, compared with 0.364 for the MAE-targeting Laplace \textsc{Forma}, a difference significant at the 1\% level. Figure \ref{fig:horizon}(b) shows a similar short-horizon pattern to the RF comparison: the GBM leads at \(h=1\) (\(32.6\%\) versus \(32.1\%\) \(R^2\); DM \(t=+6.9\)), but \textsc{Forma} leads from \(h=2\) onward: \(25.1\%\) versus \(17.2\%\) at \(h=10\) to \(19.7\%\) versus \(13.3\%\) at \(h=20\).

\paragraph{Generalists: LLMs and TSFMs.} On the task metric $R^2$, the verdict is one-sided: \textsc{Forma} achieves 29.9\% on the LLM sample, compared to 18.6\% for the top-performing LLM. In fact, every purpose-trained model other than fade/AR(1) outperforms every LLM, and every LLM outperforms Chronos-2, which beats only the seasonal random walk. The \textsc{Forma}--LLM gap widens from 5.5 percentage points at $h{=}1$ to 15.9 at $h{=}20$, with the best LLM (Opus 4.8) collapsing from 32.5\% to 7.8\% while \textsc{Forma} only decays from 38.0\% to 23.7\%. Note the comparison is conservatively biased \emph{toward} the generalists as test-period financial statement realizations sit in their pretraining data.

Panel~B supports the diagnosis promised in \S\ref{sec:eval}: the LLMs' MAEs (0.362--0.368) are only slightly higher than that of the Laplace \textsc{Forma} (\(0.348\)) and lower than those of Chronos-2 and the seasonal random walk. Consistent with conditional medians, these forecasts are competitive in absolute error but underperform under the squared-error criterion essential for valuation.

The LLM approach also has disadvantages in cost and reproducibility. Frontier models cost  $\approx$\$0.12--\$0.26 per origin while inference cost is negligible on our $\approx$0.9M-parameter \textsc{Forma}. Frontier LLMs with extended thinking produce non-deterministic outputs without the option of a reduced sampling temperature. We quantify this wobble with five repeated forecasts per origin on a 200-origin sample. For Opus 4.8, the most stable model, the cross-call standard deviation grows from 0.03 at $h{=}1$ to 0.06 at $h{=}20$ in normalized $z$-space (cross-sectional s.d.\ 1). These wobbles are independent across origins and wash out in the full 2{,}103-origin metrics.

\paragraph{Probabilistic quality.} Panel~C evaluates the models that produce predictive distributions: the specialist mixtures and Chronos-2's native quantiles; the LLMs, emitting only a point forecast, drop out. Among the mixtures, the exact five-seed NLL and closed-form CRPS rank \textsc{Forma}'s Laplace variant clearly first and its Gaussian variant second, both ahead of the FFNN variants.

Beyond the proper scores NLL and CRPS, calibration asks whether conditional quantiles and variances can be taken at face value~\cite{gneiting2007strictly}; we assess it with the mixture probability integral transform (PIT). While individual seeds are mildly overconfident (standardized MSE $\bar z^2{=}1.10$), the five-seed Gaussian mixture is essentially calibrated in total variance ($\bar z^2{=}0.96$, the between-seed spread supplies the missing variance). Its central intervals never under-cover at any horizon---pooled coverage is 72.6/89.6/93.7/95.8\% at nominal 50/80/90/95\% (per-horizon series in the release). 

The PIT is center-heavy, so the intervals are conservative rather than sharp, and by horizon $\bar z^2$ drifts from 1.00 to 0.89---the long end errs on the safe side, the right side for terminal-value use. The feed-forward mixtures are the conservative-but-blunt contrast: notably overdispersed (mixture $\bar z^2$ of 0.85 and 0.64) with worse CRPS---it is easy to be conservative, hard to be conservative \emph{and} sharp. Chronos-2 posts a competitive CRPS (0.336), with near-nominal 50\% central coverage, but still trails both \textsc{Forma} variants; its NLL is unreported because degenerate zero-width intervals on flat contexts ($\sim$0.6\% of cells) admit no usable log score even under a $10^{-4}$ scale floor.

\subsection{Coherence: Emergent and Recoverable}\label{sec:coherence}

A forecast financial statement should add up, and \textsc{Forma} largely does so without explicit constraints. Across 124.7M enforced identity instances (an identity counts at a firm-origin-horizon when every member account is forecast), the median absolute violation of \textsc{Forma}'s raw-dollar statements is 3.7\% of the identity's gross scale. The model has substantially learned accounting structure from data. Some violation is expected even from a perfect model because dollar statements plug the standardized-space conditional means into the inverse transform, and means do not commute with the nonlinear map, so part of the violation is a property of the estimand rather than prediction error.

Exact coherence is recoverable ex post. Following the forecast reconciliation tradition~\cite{wickramasuriya2019mint}, we work in raw dollars, where the identities are linear, and project each forecast statement onto the subspace where every identity holds exactly. We compare two projections: equal-weighted, which minimizes the total squared dollar adjustment, and variance-weighted, which measures each item's adjustment in units of its predictive standard deviation, mapped into dollars via the delta method, so accounts the model is less certain about absorb more of the residual; cross-account covariances are not modeled.

Table \ref{tab:coherence} shows that the variance-weighted projection drives violations to numerical zero at no statistically significant squared-error cost ($R^2$ drops by 3.8 percentage points, quarter-clustered DM $t{=}{-}1.4$) while yielding a small but significant MAE \emph{improvement} ($t{=}{+}4.8$).\footnote{The reconciled figure should not be compared to Panel~A's entries, none of which add up. A coherent-to-coherent comparison would charge every model its own reconciliation cost. Forecasters without predictive variances would need weights estimated from pre-test residuals, and those cannot be observation-specific. The MAE column scores the Gaussian conditional-mean forecasts, not Panel~B's median-targeting Laplace head, hence 0.408 vs.\ Panel~B's 0.369.} The equal-weight variant is catastrophic ($R^2$ falls to $-5.51$, MAE rises to 0.635): a dollar adjustment spread uniformly across accounts is negligible for total assets but enormous relative to small line items such as minority interest (zero at most firms), and standardized-space errors price exactly that. This contrast illustrates the setting-specific payoff of probabilistic forecasting: the variance head enables affordable reconciliation.

\begin{table}[t]
\caption{Coherence and its cost. For each enforced identity instance (firm $\times$ origin $\times$ horizon $\times$ identity where all member accounts are forecast), we calculate the absolute violation $|\sum_j s_j \hat{v}_j|$ in raw dollars as a share of the identity's gross scale $\sum_j |\hat{v}_j|$; we report the median in the identity-violations column. Reconciliation is an ex-post projection in raw-dollar space; untouched cells pass through unchanged, so all rows share the Panel~A common sample. Significance: Diebold--Mariano vs.\ the raw transformer within column (quarter-clustered, Newey--West, HLN-corrected): $^{*}/^{**}/^{***}$ raw better at 10/5/1\%, $^{\dagger}$ variant better at 1\%.}
\label{tab:coherence}
\small
\begin{tabular}{lccc}
\toprule
Variant & $R^2\!\uparrow$ & MAE$\downarrow$ & Identity viol.$\downarrow$ \\
\midrule
Transformer-only (raw)      & \textbf{0.289} & 0.408 & 3.7\% \\
\quad + variance-weighted reconciliation  & 0.251 & \textbf{0.407}$^{\dagger}$ & $\approx 0$ \\
\quad + equal-weighted reconciliation  & $-5.51$ & 0.635$^{***}$ & $\approx 0$ \\

\bottomrule
\end{tabular}
\end{table}
\subsection{Scenario Analysis by Conditioning}\label{sec:scenario}

Scenario analysis is an integral part of the valuation process whereby an analyst makes assumptions or expresses beliefs about line items such as revenues and populates the remainder of the forecast statement based on these assumptions. \textsc{Forma} natively supports this interface: any future tuple can be revealed as an input, and the remaining query tokens are forecast conditional on it. We train \textsc{Forma} on this task using the pinned-future masking objective (\S\ref{sec:masking}). Arbitrary conditioning patterns can therefore be specified at inference time without retraining.

We evaluate scenario analysis performance with an oracle experiment by pinning the \emph{true realized} revenue path (Q1--Q20) per origin. We then forecast the remaining items, and compare against the unconditional forecast on identical origins. Using the true realized revenue values for revenue assumptions makes our measurement an upper bound on scenario value. Conditional forecasting is a recognized task in the macro literature~\cite{waggoner1999conditional}, to our knowledge never posed to a learned firm-level financial statement model.

\begin{figure}[t]
\centering
\includegraphics[width=\columnwidth]{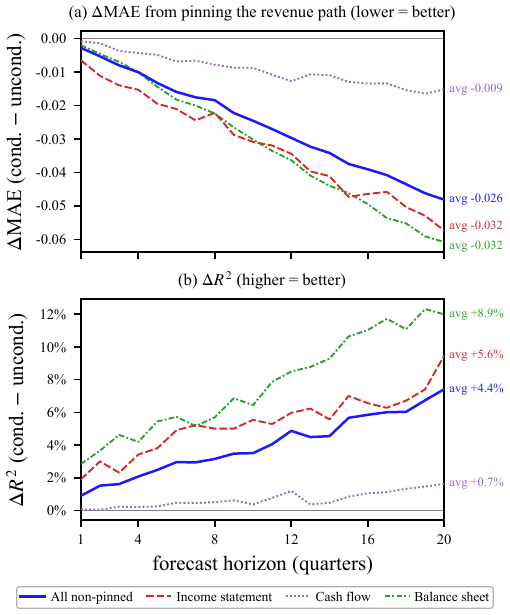}
\Description{Change in MAE and R-squared by forecast horizon from pinning the true revenue path, overall and by statement class.}
\caption{Scenario conditioning. Within-model change from pinning the true realized revenue path and forecasting the remaining statement, vs.\ the unconditional forecast on the same origins: (a)~$\Delta$MAE and (b)~$\Delta R^2$ by horizon, overall and by statement class.}
\label{fig:scenario}
\end{figure}

The results of the oracle experiment, presented in Figure \ref{fig:scenario}, confirm the interface works and that conditioning on realized revenues is economically valuable. Pinning the true revenue path lowers pooled MAE on the remaining items from 0.409 to 0.383 and raises change-space $R^2$ from 30.5\% to 34.8\% (768 origins, 916{,}846 non-pinned cells; five-seed mixture mean). The gain is small one quarter out (0.9pp) and widens with horizon to 7.4pp by $h{=}20$. Near-term items are already determined, but at long horizons the revenue path anchors the whole statement. Income-statement and balance-sheet items gain equally in absolute error ($\Delta$MAE $-0.032$ each). Balance-sheet items gain most in $R^2$ ($+8.9$pp vs.\ $+5.6$pp). Cash-flow items barely move ($\Delta R^2$ $+0.7$pp), reflecting their weak link to revenue.

\section{Conclusion}\label{sec:conclusion}

Forecasting complete financial statements at valuation-relevant horizons is a task that, to our knowledge, no prior work covers. We show that it is tractable: \textsc{Forma}, a sub-million-parameter transformer with a missingness-native representation and an identity-aware masked objective, beats classical ML, the statement-forecasting system of \citet{geertsema2026chained}, a zero-shot foundation model, and frontier LLMs. \textsc{Forma}'s lead widens toward the horizons where firm value actually sits, and its Gaussian prediction intervals never fall below nominal coverage.

Some limitations apply. Long-horizon evaluation conditions on realized survivors: comparisons are fair (identical cells for all models), but absolute skill levels describe only the survivors. Our claims are scoped to quarterly U.S. filings, these 78 accounting items, and horizons of 1--20 quarters. We measure forecast quality only; valuation applications require additional inferences about discount rates and terminal values. We do not incorporate stock market or analyst data as additional features, and only consider off-the-shelf generalists rather than fine-tuned ones.

We release \benchname{} for adoption as a common task framework for statement forecasting~\cite{hellum2025ctf}: pipeline, configurations, and evaluation protocol are rebuildable from WRDS with one command (\pipelineurl). We also release code, configurations, and trained weights for \textsc{Forma}, and seeded regeneration scripts for the learned competitors (\releaseurl). The \apxnoun{} provides additional details.

\bibliographystyle{plainnat}
\bibliography{refs}

\begingroup
  \let\origtabular\tabular
  \let\endorigtabular\endtabular
  \renewenvironment{tabular}[1]%
    {\begin{adjustbox}{max width=\linewidth}\begin{origtabular}{#1}}%
    {\end{origtabular}\end{adjustbox}}
  \let\origtabularx\tabularx
  \let\endorigtabularx\endtabularx
  \renewenvironment{tabularx}[2]%
    {\let\tabular\origtabular \let\endtabular\endorigtabular
     \origtabularx{#1}{#2}}%
    {\endorigtabularx}
  \singlespacing
  \makeatletter\setlength{\@fptop}{0pt}\makeatother
  \let\origincludegraphics\includegraphics
  \renewcommand{\includegraphics}[2][]{%
    \origincludegraphics[#1,max totalheight=0.80\textheight]{#2}}
  \clearpage
  \ifdefined\pdfpageattr          
    \pdfpageattr{/Rotate 90}%
    \gdef\unrotatepage{\global\pdfpageattr{}}%
  \else\ifdefined\pdfvariable     
    \pdfvariable pageattr{/Rotate 90}%
    \gdef\unrotatepage{\global\pdfvariable pageattr{}}%
  \else                           
    \gdef\unrotatepage{}%
  \fi\fi
  \let\origefloatseparator\efloatseparator
  \renewcommand{\efloatseparator}{%
    \origefloatseparator
    \unrotatepage
    \global\let\efloatseparator\origefloatseparator}%
  \processdelayedfloats
\endgroup

\makeatletter
\renewenvironment{table}{\@float{table}}{\end@float}
\makeatother

\apxstart
\titleformat{\section}[hang]{\normalfont\centering\bfseries}%
  {\appendixname~\thesection.}{10pt}{}{}
\singlespacing

\section{Data}\label{app:data}

This \apxnoun{} documents the data underlying the benchmark: the pipeline from the raw Compustat pull to the modeled firm-quarter origins (\S\ref{app:pipeline}), the standardization that maps reported accounting values into model space (\S\ref{app:transformation}), the 78-item target universe (\S\ref{app:universe}), its reporting availability (\S\ref{app:sparsity}), and the accounting identities that link the items (\S\ref{app:identities}).

Throughout, a \emph{forecast origin} is a firm-quarter $(f,t)$ from which a model forecasts quarters $t{+}1,\dots,t{+}20$ using the 12 quarters through the origin ($t{-}11,\dots,t$). The benchmark provides the same panel in two views: the \emph{tuple view}, in which each firm-origin is a variable-length set of (relative quarter, item, value) tuples, consumed by \textsc{Forma} and the LLM protocol, and the \emph{tabular view}, a fixed-width matrix of recent levels and year-over-year changes per item, consumed by the tabular baselines (\S\ref{app:competitors}).

\subsection{Sample formation and splits}\label{app:pipeline}

Table~\ref{tab:app-waterfall} gives the sample waterfall; the paragraphs below document each stage in order. The scale deflator that defines the final filter is specified in \S\ref{app:transformation}.

\begin{table}[htbp]
\centering\small
\caption{Sample formation waterfall. Row 1 is the raw \texttt{comp.fundq} pull and each subsequent row applies one filter or transform to the row above. Row 1 is a WRDS re-query at a marginally later vintage than the downloaded panel, so the implied link-window trim is approximate (see text). The last waterfall row is the deflator-valid firm-quarter panel whose counts the main paper quotes (dropping the 203 deflator-valid firm-quarters dated 1970, which serve only as history); the rows below the rule partition it by split. Origin \emph{eligibility} additionally requires four quarters of history through the origin (see text).}
\label{tab:app-waterfall}
\begin{tabular}{lr}
\toprule
Stage & Firm-quarters \\
\midrule
Raw Compustat quarterly rows (1970--2024 pull) & 2,002,987 \\
After CRSP link-window trim (see text) & 1,708,447 \\
Distinct firm-quarters (after deduplication) & 1,706,140 \\
After financial-sector exclusion (SIC 6000--6999) & 1,531,223 \\
With a valid scale deflator & 1,173,801 \\
Deflator-valid firm-quarters (1971--2024 window) & 1,173,598 \\
\midrule
\quad Train (1971--2001) & 609,269 \\
\quad Validation (2002--2009) & 211,367 \\
\quad Test (2010--2024) & 352,962 \\
\bottomrule
\end{tabular}
\end{table}

\paragraph{Download and universe.} The raw panel is Compustat Fundamentals Quarterly (\texttt{comp.fundq}) via WRDS, restricted to \texttt{indfmt=INDL}, \texttt{datafmt=STD}, \texttt{consol=C}, \texttt{popsrc=D}, with \texttt{datadate} in 1970--2024. Historical SIC codes (\texttt{sich}) come from \texttt{comp.co\_industry}, attached to each firm-quarter as the most recent industry record at or before the statement date. The financial-sector filter drops firm-quarters with $\texttt{sich}\in[6000,6999]$; firm-quarters with \emph{missing} SIC are retained and map to the Unknown industry bucket.

The download also intersects the panel with the CRSP--Compustat link table (\texttt{crsp.ccmxpf\_lnkhist}, link types LU/LC, primary flags P/C): firms that appear in the link table retain only firm-quarters whose statement date falls within a link window (dropping quarters before listing and after delisting), while the 16{,}418 firms with no link-table entry are retained in full (495{,}440 firm-quarters, 29\% of the retained panel). The trim removes 294{,}540 firm-quarters, 15\% of the raw pull, and leaves 26{,}917 link-matched firms alongside the 16{,}418 unmatched ones. Because WRDS updates \texttt{comp.fundq} and the link table in place, the raw count in Table~\ref{tab:app-waterfall} is a marginally later vintage than the downloaded panel; replaying the trim entirely on the re-queried vintage drops 296{,}589 firm-quarters, about 0.1\% of the raw pull --- 0.7\% of the trim itself --- away from the 294{,}540 implied here.

\paragraph{Deduplication and quarter alignment.} Firm-quarters are indexed by \emph{calendar} quarter: each row's quarter is the calendar quarter-end of its \texttt{datadate}. When multiple rows share a (firm, calendar-quarter) pair (e.g.\ around fiscal-year changes), rows are sorted by fiscal year and quarter and the last record is kept. Year-to-date accumulation, by contrast, follows the \emph{fiscal} calendar (next paragraph), so flow conversion is unaffected by fiscal/calendar offsets.

\paragraph{Year-to-date conversion.} Compustat reports the 20 cash-flow items year-to-date. Each is converted to a quarterly flow within the firm--fiscal-year group: fiscal Q1 values pass through unchanged, and fiscal Q2--Q4 values are differenced against the immediately preceding fiscal quarter's year-to-date value. If that preceding fiscal quarter is not in the file, the quarterly flow is left missing rather than imputed; no differencing ever crosses a fiscal-year boundary.

\paragraph{Derived items.} The six derived items are computed after year-to-date conversion, exactly as defined in Table~\ref{tab:app-dictionary}: gross profit ($\texttt{gpq}=\texttt{revtq}-\texttt{cogsq}$), free cash flow ($\texttt{fcfq}=\texttt{oancfq}-\texttt{capxq}$), working capital ($\texttt{wcapq}=\texttt{actq}-\texttt{lctq}$), and the three carve-outs $\texttt{aoq\_ex\_intanq}$, $\texttt{loq\_ex\_dr}$, $\texttt{xsgaq\_ex\_rd}$, in which the subtracted component (\texttt{intanq}, \texttt{drltq}, \texttt{xrdq}) is treated as zero when unreported.

\paragraph{Origin eligibility and splits.} A firm-quarter is an eligible forecast origin if it has a valid deflator and at least four quarters of history through the origin. The history requirement is enforced per firm at load time, so a firm's first three panel quarters cannot serve as origins; the tabular build instead drops the panel's first three calendar quarters globally. The two rules differ only at firm entry, and scored comparisons are unaffected because every comparison is made on the common sample of cells all compared models predict. The counts in Table~\ref{tab:app-waterfall} are the deflator-valid firm-quarter panel \emph{before} this final gate; eligible-origin counts are smaller and depend on a consumer's additional requirements (e.g.\ the LLM protocol's test pool of 321{,}892 origins, \S\ref{app:llm-protocol}). Every input window spans the trailing 12 quarters, with unreported (item, quarter) cells contributing no tuple. Splits are by the calendar year of the origin quarter: 1971--2001 train, 2002--2009 validation, 2010--2024 test. Split boundaries also purge targets: a train (validation) example retains only target quarters dated on or before 2001Q4 (2009Q4) --- later targets are set to missing in the tabular view and are absent from the split's tuple file --- so origins near a boundary contribute only their realized short horizons and no target crosses a split boundary. Hyperparameter and model-selection decisions for every model use the validation split only, whose targets end in 2009Q4; final models are refit on train$+$validation with all choices frozen (\S\ref{app:forma}, \S\ref{app:competitors}), so nothing trained or selected before test conditions on an outcome after 2009Q4.

\subsection{Standardization}\label{app:transformation}

Reported values span orders of magnitude across firms and are heavy-tailed within firm. The pipeline therefore maps every accounting value into a standardized space before it reaches any model, in four steps: deflation by firm size, an asinh transform, cross-sectional standardization, and clipping. We call the full mapping from an original value $v$ to a model value $x$ \emph{standardization}, and its inverse \emph{de-standardization}. All standardization parameters are fixed at the forecast origin. The same procedure serves the tuple view and the tabular view, and forecasts are evaluated in the standardized space.

\paragraph{Scale deflator.} Let $v_{f,q,\mathrm{id}}$ denote the reported value of item $\mathrm{id}$ for firm $f$ at quarter $q$. The deflator of firm-quarter $(f,q)$ is
\[
z_{f,q}
\equiv
\begin{cases}
\left|v_{f,q,\texttt{ltq}}\right| + \left|v_{f,q,\texttt{seqq}}\right| + 10^{-3}
& \text{when valid},\\[4pt]
\left|v_{f,q,\texttt{atq}}\right| + 10^{-3}
& \text{otherwise},
\end{cases}
\]
i.e.\ total liabilities plus total stockholders' equity, which by the balance identity (\S\ref{app:identities}) equals total assets less noncontrolling interests, up to sign conventions. The primary definition is valid when both components are reported and the sum is finite and positive; the total-assets fallback rescues 6{,}350 firm-quarters, and firm-quarters with no valid deflator under either definition are dropped from the panel (357{,}422 firm-quarters; Table~\ref{tab:app-waterfall}). We call the surviving firm-quarters \emph{scale-valid}. For an origin $(f,t)$, every model input and target --- at every lead and lag in the window --- is deflated by the origin deflator $z_{f,t}$; deflators at other quarters enter only the normalization statistics below.

\paragraph{Scaling and transform.} For a firm $f$ with forecast origin $t$, the scaled value of an accounting tuple $(h,\mathrm{id},v)$ --- item $\mathrm{id}$ reporting value $v$ at relative quarter $h$, i.e.\ absolute quarter $t+h$ --- is
\[
v^{\mathrm{scaled}}_{h,\mathrm{id}}(f,t)
\equiv
\operatorname{asinh}\!\left(
k_{\mathrm{id}}\,
\frac{v}{z_{f,t}}
\right),
\]
so the entire example, leads and lags alike, is expressed on one origin-fixed scale. The account-specific constant $k_{\mathrm{id}}$ is estimated once on training-period observations (quarters through 2001Q4) of the lagged-ratio pool defined for $\sigma$ below, and is then held fixed: it is the value on a 250-point logarithmic grid over $[10^{-2},10^{3}]$ that brings the excess kurtosis of the pooled transformed values as close as possible to~3 (total kurtosis~6, a mildly heavier-tailed target than the Gaussian).

\paragraph{Cross-sectional standardization.} Scaled values are standardized per (item, quarter) with statistics built from trailing cross-sections. Let $\mathcal{F}$ denote the cross-sectional pool of firms. For each calendar quarter $q$, define the raw per-quarter statistics
\[
\mu^{\mathrm{raw}}_{\mathrm{id},q}
\equiv
\operatorname{mean}_{f'\in\mathcal{F}}
\operatorname{asinh}\!\left(
k_{\mathrm{id}}\,
\frac{v_{f',q,\mathrm{id}}}{z_{f',q}}
\right),
\qquad
\sigma^{\mathrm{raw}}_{\mathrm{id},q}
\equiv
\operatorname{std}_{\substack{
f'\in\mathcal{F},\;
r\in\{0,-8\}
}}
\operatorname{asinh}\!\left(
k_{\mathrm{id}}\,
\frac{v_{f',q+r,\mathrm{id}}}{z_{f',q-4}}
\right).
\]
The mean is the contemporaneous cross-section of item $\mathrm{id}$ at quarter $q$, each observation deflated by its own quarter's scale. The standard deviation is computed over the pooled values at quarters $q$ and $q-8$, both deflated by the scale of the intermediate quarter $q-4$ (the lagged-ratio pool). In both constructions a lag indexes positions in the firm's reported series for the item, so reporting gaps collapse. The origin-quarter parameters average the four most recent raw statistics (fewer at the panel start),
\[
\mu_{\mathrm{id},t}
\equiv
\frac{1}{4}
\sum_{q=t-3}^{t}
\mu^{\mathrm{raw}}_{\mathrm{id},q},
\qquad
\sigma_{\mathrm{id},t}
\equiv
\frac{1}{4}
\sum_{q=t-3}^{t}
\sigma^{\mathrm{raw}}_{\mathrm{id},q},
\]
and the normalized value is
\[
v^{\mathrm{normalized}}_{h,\mathrm{id}}(f,t)
\equiv
\frac{
v^{\mathrm{scaled}}_{h,\mathrm{id}}(f,t)
-
\mu_{\mathrm{id},t}
}{
\sigma_{\mathrm{id},t}+10^{-8}
},
\]
with $10^{-8}$ a numerical floor. All means and standard deviations use available reported observations only. Every lag in the construction is non-negative and the four-quarter averages are trailing, so $(\mu_{\mathrm{id},t},\sigma_{\mathrm{id},t})$ depend only on data through quarter $t$ and are updated each quarter; the per-item constants $k_{\mathrm{id}}$ are the only standardization parameters frozen at the 2001Q4 estimation cutoff.

\paragraph{Clipping.} With
$\operatorname{clip}(u,a,b)\equiv\min\{b,\max\{a,u\}\}$, the model value of a
tuple $(h,\mathrm{id},v)$ is
\[
x_{f,t}(h,\mathrm{id})
\equiv
\operatorname{standardize}_{f,t}(h,\mathrm{id},v)
\equiv
\operatorname{clip}\!\left(
v^{\mathrm{normalized}}_{h,\mathrm{id}}(f,t),
-6,6
\right).
\]
The same clipped value serves as a model input when the tuple is visible and as the supervised target when it is masked. In the tuple view the clip is applied at load time; in the tabular view it is applied at build time, to levels and targets at $6$ and to year-over-year difference features, which are computed from the unclipped standardized levels, at $6\sqrt{2}$ (the $\pm 6$ bound scaled by $\sqrt{2}$, the standard deviation of the difference of two independent unit-variance values). Because clipping is not invertible, an original value whose standardized value lies at either boundary cannot generally be recovered uniquely.

\paragraph{The scale token.} In the tuple view the deflator itself enters as one additional input tuple per origin, under a dedicated $\mathrm{scale}$ account at $h=0$, so models observe firm size. Its value is $z_{f,t}$, standardized like any other account through an asinh transform with its own constant $k_{\mathrm{scale}}$ and per-quarter statistics $(\mu_{\mathrm{scale},t},\sigma_{\mathrm{scale},t})$ estimated on the cross-section of contemporaneous deflator values; the deflator is not divided by itself. The scale account is never queried at future horizons and receives no future-grid slots (\S\ref{app:forma}); like any historical tuple, it can be masked for reconstruction during training. In the tabular view the deflator enters as one feature on the same standardized basis.

\paragraph{De-standardization.} Given a predicted model value $\widehat{x}_{f,t}(h,\mathrm{id})$, the prediction in original accounting units is
\[
\widehat{v}_{f,t+h,\mathrm{id}}
\equiv
z_{f,t}\,
\frac{
\sinh\!\left(
\widehat{x}_{f,t}(h,\mathrm{id})\,
\sigma_{\mathrm{id},t}
+
\mu_{\mathrm{id},t}
\right)
}{
k_{\mathrm{id}}
}
\]
(numerical floors of order $10^{-8}$ omitted). De-standardization uses the deflator and normalization parameters fixed at the forecast origin $t$, never the realization quarter $t+h$. The map is strictly increasing in $\widehat{x}$, so it carries quantiles of a predictive distribution in model space to quantiles in accounting units, a property the Chronos-2 comparison relies on (\S\ref{app:competitors}).

\subsection{The \benchname{} variable universe}\label{app:universe}

\benchname{} targets the 78 quarterly Compustat line items in Table~\ref{tab:app-dictionary}: 34 balance-sheet, 18 income-statement, 20 cash-flow, and 6 derived items, jointly at horizons $h=1,\dots,20$. The same items serve as both conditioning inputs (reported history) and prediction targets (future quarters); \benchname{} additionally conditions on a Fama--French-48 industry category. Cash-flow items are converted from year-to-date to quarterly flows (\S\ref{app:pipeline}); derived items are exact functions of raw items (definitions in the table). Mnemonics are Compustat quarterly field names.

{\footnotesize
\setlength{\tabcolsep}{4pt}
\begin{longtable}{llcrrrr}
\caption{The 78-item \benchname{} target universe, ordered alphabetically by Compustat mnemonic: description, statement (St: BS balance sheet, IS income statement, CF cash flow, Der derived), and reporting availability. \emph{\%\,raw} is the fraction of all 1{,}706{,}140 Compustat quarterly firm-quarters (all sectors and years, no scale-validity requirement) that report the item after year-to-date$\rightarrow$quarterly conversion, i.e.\ the reporting rate before any sample filter. The split columns are that same fraction restricted to the scale-valid firm-quarters of each split: train 1971--2001 ($N{=}609{,}269$), validation 2002--2009 ($N{=}211{,}367$), test 2010--2024 ($N{=}352{,}962$). Items near-zero in the training era (e.g.\ \texttt{stkcoq}, \texttt{txbcofq}, \texttt{drcq}, \texttt{acomincq}) are modern disclosure fields absent from early filings; \textsc{Forma} represents them as missing rather than imputing them.}\label{tab:app-dictionary}\\
\toprule
Mnemonic & Description & St & \%\,raw & \%\,Tr & \%\,Val & \%\,Te \\
\midrule
\endfirsthead
\multicolumn{7}{l}{\footnotesize\itshape Table~\ref{tab:app-dictionary}, continued}\\
\toprule
Mnemonic & Description & St & \%\,raw & \%\,Tr & \%\,Val & \%\,Te \\
\midrule
\endhead
\midrule
\multicolumn{7}{r}{\footnotesize\itshape continued on next page}\\
\endfoot
\bottomrule
\endlastfoot
\texttt{acomincq} & Accumulated other comprehensive income & BS & 38.7 & 1.5 & 92.3 & 98.0 \\
\texttt{acoq} & Current assets, other & BS & 72.2 & 92.1 & 96.6 & 99.7 \\
\texttt{actq} & Current assets, total & BS & 65.5 & 89.6 & 94.0 & 97.5 \\
\texttt{ancq} & Non-current assets, total & BS & 62.6 & 85.0 & 90.3 & 93.8 \\
\texttt{aoq} & Assets, other & BS & 78.1 & 98.4 & 99.8 & 99.9 \\
\texttt{aoq\_ex\_intanq} & Other assets excl. intangibles & Der & 78.1 & 98.4 & 99.8 & 99.9 \\
\texttt{apq} & Accounts payable & BS & 76.9 & 96.8 & 98.9 & 99.3 \\
\texttt{aqcq} & Acquisitions & CF & 61.9 & 64.8 & 90.2 & 91.2 \\
\texttt{atq} & Assets, total & BS & 78.8 & 100.0 & 100.0 & 100.0 \\
\texttt{capsq} & Capital surplus / APIC & BS & 75.8 & 93.6 & 95.1 & 94.9 \\
\texttt{capxq} & Capital expenditures & CF & 62.7 & 65.1 & 91.9 & 93.7 \\
\texttt{ceqq} & Common equity, total & BS & 79.8 & 98.8 & 99.5 & 99.8 \\
\texttt{cheq} & Cash \& short-term investments & BS & 77.8 & 97.8 & 99.8 & 99.9 \\
\texttt{cogsq} & Cost of goods sold & IS & 79.7 & 97.5 & 98.9 & 99.5 \\
\texttt{cstkq} & Common stock & BS & 76.7 & 95.0 & 95.3 & 96.2 \\
\texttt{dlcchq} & Change in current debt & CF & 34.0 & 32.6 & 51.7 & 54.9 \\
\texttt{dlcq} & Debt in current liabilities & BS & 75.2 & 93.8 & 97.4 & 97.5 \\
\texttt{dltisq} & Long-term debt issuance & CF & 60.6 & 62.7 & 87.4 & 91.8 \\
\texttt{dltrq} & Long-term debt reduction & CF & 61.1 & 62.9 & 88.9 & 92.3 \\
\texttt{dlttq} & Long-term debt & BS & 79.5 & 98.6 & 99.2 & 99.3 \\
\texttt{dpactq} & Accumulated depreciation & BS & 51.1 & 73.2 & 57.9 & 69.6 \\
\texttt{dpq} & Depreciation \& amortization & IS & 69.5 & 81.9 & 92.8 & 96.2 \\
\texttt{drcq} & Deferred revenue, current & BS & 29.5 & 0.1 & 67.4 & 89.3 \\
\texttt{drltq} & Deferred revenue, long-term & BS & 31.3 & 0.1 & 69.9 & 92.6 \\
\texttt{dvq} & Cash dividends & CF & 63.1 & 66.1 & 91.4 & 93.4 \\
\texttt{exreq} & Exchange rate effect on cash & CF & 57.0 & 49.1 & 92.3 & 94.1 \\
\texttt{fcfq} & Free cash flow $=$ oancfq $-$ capxq & Der & 56.3 & 48.0 & 91.9 & 93.7 \\
\texttt{fiaoq} & Financing activities, other & CF & 56.9 & 49.1 & 92.1 & 93.8 \\
\texttt{fincfq} & Financing cash flow & CF & 57.1 & 49.3 & 92.5 & 94.3 \\
\texttt{fopoq} & Funds from operations, other & CF & 60.1 & 59.7 & 89.7 & 93.0 \\
\texttt{gdwlq} & Goodwill & BS & 39.3 & 4.2 & 90.9 & 97.1 \\
\texttt{gpq} & Gross profit $=$ revtq $-$ cogsq & Der & 74.0 & 92.0 & 95.7 & 99.2 \\
\texttt{ibq} & Income before extraordinary items & IS & 82.6 & 99.6 & 99.5 & 99.7 \\
\texttt{intanoq} & Intangibles, other & BS & 37.8 & 4.3 & 87.4 & 92.6 \\
\texttt{intanq} & Intangibles, total & BS & 43.0 & 8.9 & 99.0 & 99.5 \\
\texttt{invtq} & Inventory & BS & 75.9 & 95.0 & 97.7 & 98.1 \\
\texttt{ivacoq} & Investing activities, other & CF & 56.9 & 49.1 & 92.2 & 93.9 \\
\texttt{ivchq} & Increase in investments & CF & 60.7 & 63.3 & 88.2 & 90.6 \\
\texttt{ivncfq} & Investing cash flow & CF & 57.1 & 49.3 & 92.5 & 94.3 \\
\texttt{ivstchq} & Short-term investments, change & CF & 45.7 & 45.0 & 77.1 & 73.5 \\
\texttt{lcoq} & Current liabilities, other & BS & 72.0 & 91.5 & 96.4 & 99.6 \\
\texttt{lctq} & Current liabilities, total & BS & 66.0 & 90.2 & 94.1 & 97.6 \\
\texttt{loq} & Liabilities, other & BS & 78.1 & 98.3 & 99.8 & 99.9 \\
\texttt{loq\_ex\_dr} & Other liab. excl. deferred rev. & Der & 78.1 & 98.3 & 99.8 & 99.9 \\
\texttt{ltq} & Liabilities, total & BS & 78.4 & 99.1 & 99.9 & 99.9 \\
\texttt{mibtq} & Noncontrolling interests (BS) & BS & 74.5 & 91.5 & 93.7 & 98.1 \\
\texttt{miiq} & Noncontrolling interest (income) & IS & 76.2 & 88.9 & 91.8 & 96.9 \\
\texttt{niq} & Net income & IS & 82.6 & 99.6 & 99.4 & 99.7 \\
\texttt{nopiq} & Non-operating income & IS & 79.9 & 98.2 & 98.9 & 99.5 \\
\texttt{oancfq} & Operating cash flow & CF & 57.1 & 49.3 & 92.5 & 94.3 \\
\texttt{oiadpq} & Operating income after D\&A (EBIT) & IS & 79.0 & 96.9 & 98.4 & 99.2 \\
\texttt{oibdpq} & Operating income before D\&A (EBITDA) & IS & 70.6 & 83.3 & 93.2 & 96.1 \\
\texttt{piq} & Pretax income & IS & 81.8 & 98.5 & 99.3 & 99.7 \\
\texttt{ppegtq} & Gross PP\&E & BS & 51.0 & 73.2 & 57.9 & 69.6 \\
\texttt{ppentq} & Net PP\&E & BS & 76.6 & 99.2 & 99.5 & 99.6 \\
\texttt{prstkcq} & Purchase of common \& preferred stock & CF & 59.9 & 62.4 & 85.2 & 89.5 \\
\texttt{pstkq} & Preferred stock & BS & 79.8 & 98.8 & 99.4 & 99.7 \\
\texttt{rectq} & Receivables, total & BS & 75.4 & 94.3 & 97.6 & 97.4 \\
\texttt{req} & Retained earnings & BS & 76.3 & 94.5 & 95.7 & 95.3 \\
\texttt{revtq} & Revenue, total & IS & 76.8 & 93.8 & 96.1 & 99.4 \\
\texttt{seqq} & Stockholders' equity, total & BS & 80.3 & 99.9 & 99.9 & 99.9 \\
\texttt{sivq} & Sale of investments & CF & 60.8 & 63.2 & 88.6 & 91.0 \\
\texttt{spiq} & Special items & IS & 77.1 & 89.8 & 98.8 & 98.3 \\
\texttt{sppeq} & Sale of PP\&E & CF & 52.0 & 52.3 & 76.0 & 79.8 \\
\texttt{sstkq} & Sale of common \& preferred stock & CF & 62.3 & 64.9 & 90.7 & 92.4 \\
\texttt{stkcoq} & Stock compensation expense & IS & 30.1 & 0.1 & 62.5 & 83.9 \\
\texttt{tstkq} & Treasury stock & BS & 74.5 & 86.5 & 98.9 & 99.1 \\
\texttt{txbcofq} & Excess tax benefit, stock options & CF & 30.8 & 0.0 & 49.0 & 93.5 \\
\texttt{txditcq} & Deferred taxes \& investment tax credit & BS & 68.5 & 87.8 & 91.3 & 93.4 \\
\texttt{txpq} & Income taxes payable & BS & 66.5 & 84.1 & 88.2 & 92.7 \\
\texttt{txtq} & Income taxes, total & IS & 81.9 & 98.6 & 99.3 & 99.7 \\
\texttt{wcapq} & Working capital $=$ actq $-$ lctq & Der & 65.4 & 89.4 & 93.9 & 97.5 \\
\texttt{xidoq} & Extraordinary items \& disc. ops & IS & 82.6 & 99.5 & 99.4 & 99.7 \\
\texttt{xintq} & Interest expense & IS & 65.8 & 79.0 & 81.4 & 88.6 \\
\texttt{xoprq} & Operating expenses, total & IS & 79.5 & 97.2 & 98.8 & 99.3 \\
\texttt{xrdq} & R\&D expense & IS & 24.2 & 22.3 & 41.4 & 43.6 \\
\texttt{xsgaq} & SG\&A expense & IS & 62.6 & 77.9 & 83.4 & 85.2 \\
\texttt{xsgaq\_ex\_rd} & SG\&A excl. R\&D & Der & 62.6 & 77.9 & 83.4 & 85.2 \\
\end{longtable}
}

\subsection{Availability and sparsity}\label{app:sparsity}

Reporting is pervasively incomplete. Table~\ref{tab:app-joint} quantifies the joint availability of the 78-item vector across scale-valid firm-quarters. On the test panel the item$\times$firm-quarter matrix is 93.6\% dense and the average firm-quarter reports 73 of 78 items, yet only 4.7\% report all 78, so complete-case (listwise) deletion, the implicit requirement of a dense tabular design, would discard 95.3\% of the test panel. \textsc{Forma}'s tuple-set representation accommodates this incompleteness natively; the tabular baselines require an explicit imputation model.

\begin{table}[htbp]
\centering\small
\caption{Joint availability of the 78-item vector across scale-valid firm-quarters. \emph{Matrix density} is the fraction of item$\times$firm-quarter cells reported; \emph{items per firm-quarter} summarizes how many of the 78 items a firm reports in a given quarter; \emph{complete-case} firm-quarters report all 78.}
\label{tab:app-joint}
\begin{tabular}{lrr}
\toprule
 & Full sample & Test \\
 & 1971--2024 & 2010--2024 \\
\midrule
Scale-valid firm-quarters & 1,173,598 & 352,962 \\
Distinct firms & 32,851 & 13,288 \\
Matrix density (\%) & 82.2 & 93.6 \\
Items per firm-quarter, mean & 64.1 & 73.0 \\
\quad median\,[p10,\,p90] & 68\,[46,\,76] & 75\,[68,\,77] \\
With $\ge 40$ items (\%) & 95.1 & 99.7 \\
With $\ge 70$ items (\%) & 41.4 & 87.1 \\
Complete-case (all 78, \%) & 1.73 & 4.67 \\
\midrule
\multicolumn{3}{l}{\textit{Cell density by statement (\%), full / test}}\\
\quad Balance sheet & 85.4 & 95.8 \\
\quad Income statement & 87.8 & 93.6 \\
\quad Cash flow & 69.7 & 89.3 \\
\quad Derived & 89.4 & 95.9 \\
\bottomrule
\end{tabular}
\end{table}

\subsection{Accounting identities}\label{app:identities}

The 26 exact linear identities below tie the target items. They define the identity groups used by \textsc{Forma}'s grouped masking objective and the coherence residuals analyzed in the main paper. Each holds in levels at every quarter; contemporaneous subscripts are suppressed. Signs are normalized so all terms are additive (negative-coefficient terms moved to the right-hand side).

\paragraph{Balance sheet (13).}
\begin{align*}
\texttt{cheq} + \texttt{invtq} + \texttt{rectq} + \texttt{acoq} + \texttt{ppentq} + \texttt{aoq} &= \texttt{atq} \\
\texttt{actq} + \texttt{ancq} &= \texttt{atq} \\
\texttt{cheq} + \texttt{invtq} + \texttt{rectq} + \texttt{acoq} &= \texttt{actq} \\
\texttt{wcapq} + \texttt{lctq} &= \texttt{actq} \\
\texttt{intanq} + \texttt{aoq\_ex\_intanq} &= \texttt{aoq} \\
\texttt{gdwlq} + \texttt{intanoq} &= \texttt{intanq} \\
\texttt{ppentq} + \texttt{dpactq} &= \texttt{ppegtq} \\
\texttt{apq} + \texttt{lcoq} + \texttt{dlcq} + \texttt{txpq} + \texttt{dlttq} + \texttt{txditcq} + \texttt{loq} &= \texttt{ltq} \\
\texttt{apq} + \texttt{dlcq} + \texttt{txpq} + \texttt{lcoq} &= \texttt{lctq} \\
\texttt{drltq} + \texttt{loq\_ex\_dr} &= \texttt{loq} \\
\texttt{cstkq} + \texttt{capsq} + \texttt{req} &= \texttt{tstkq} + \texttt{ceqq} \\
\texttt{seqq} &= \texttt{pstkq} + \texttt{ceqq} \\
\texttt{atq} &= \texttt{ltq} + \texttt{mibtq} + \texttt{seqq}
\end{align*}
\paragraph{Income statement (10).}
\begin{align*}
\texttt{gpq} + \texttt{cogsq} &= \texttt{revtq} \\
\texttt{oibdpq} + \texttt{xsgaq} &= \texttt{gpq} \\
\texttt{xrdq} + \texttt{xsgaq\_ex\_rd} &= \texttt{xsgaq} \\
\texttt{xoprq} &= \texttt{cogsq} + \texttt{xsgaq} \\
\texttt{oiadpq} + \texttt{dpq} &= \texttt{oibdpq} \\
\texttt{piq} + \texttt{xintq} &= \texttt{oiadpq} + \texttt{nopiq} + \texttt{spiq} \\
\texttt{ibq} + \texttt{txtq} + \texttt{miiq} &= \texttt{piq} \\
\texttt{niq} &= \texttt{ibq} + \texttt{xidoq} \\
\texttt{niq} + \texttt{txtq} + \texttt{miiq} &= \texttt{piq} + \texttt{xidoq} \\
\texttt{revtq} + \texttt{nopiq} + \texttt{spiq} + \texttt{xidoq} &= \texttt{cogsq} + \texttt{xsgaq} + \texttt{dpq} + \texttt{xintq} + \texttt{txtq} + \texttt{miiq} + \texttt{niq}
\end{align*}
\paragraph{Cash flow (3).}
\begin{align*}
\texttt{sivq} + \texttt{sppeq} + \texttt{ivstchq} + \texttt{ivacoq} &= \texttt{capxq} + \texttt{ivchq} + \texttt{aqcq} + \texttt{ivncfq} \\
\texttt{sstkq} + \texttt{dltisq} + \texttt{dlcchq} + \texttt{fiaoq} + \texttt{txbcofq} &= \texttt{prstkcq} + \texttt{dltrq} + \texttt{dvq} + \texttt{fincfq} \\
\texttt{fcfq} + \texttt{capxq} &= \texttt{oancfq}
\end{align*}

\clearpage

\section{Models}\label{app:models}

The main paper specifies \textsc{Forma}'s representation and objective. This \apxnoun{} records the implementation detail needed to retrain it from scratch (\S\ref{app:forma}) and the exact specification of every competitor (\S\ref{app:competitors}). The released configuration files are the authoritative source for all values quoted here. Throughout, the \emph{headline table} is the main paper's headline accuracy table, which scores three tracks on standardized values: a squared-error track summarized by change-space $R^2$ (the $R^2$ of predicted against realized changes $x_{t+h}-x_t$, so zero corresponds to predicting the sample-average change), an absolute-error track (MAE), and a density track (log score, CRPS, and coverage) for models that produce predictive distributions.

\subsection{\textsc{Forma}: training and implementation}\label{app:forma}

Table~\ref{tab:app-forma-hparams} collects the canonical hyperparameters.

\begin{table}[htbp]
\centering\small
\caption{\textsc{Forma} canonical hyperparameters (identical across the five
mixture seeds; only the seed differs).}
\label{tab:app-forma-hparams}
\begin{tabular}{ll}
\toprule
\multicolumn{2}{l}{\textit{Architecture}}\\
Encoder layers / $d_{\mathrm{model}}$ / heads & 4 / 128 / 4 \\
Feed-forward width / dropout & 512 / 0.2 \\
Trainable parameters (incl.\ variance head) & 942{,}210 \\
\midrule
\multicolumn{2}{l}{\textit{Optimization}}\\
Optimizer & AdamW (default $\beta$, $\epsilon$) \\
Learning rate & $10^{-4}$, constant \\
Weight decay & 0.1 (all parameters) \\
Gradient-norm clip & 1.0 \\
Batch size & 32 firm-origin sets \\
Epochs & 12 (final-epoch weights used) \\
Precision & fp32 (\texttt{medium} matmul precision) \\
Seeds & 60--64 (equal-weight mixture) \\
\midrule
\multicolumn{2}{l}{\textit{Loss}}\\
Objective & Gaussian $\beta$-NLL, $\beta=0.5$ \\
$\log\sigma^2$ clamp & $[-10, 10]$ \\
Absolute-error-track variant & Laplace $\beta$-NLL, $b=\sigma/\sqrt{2}$ \\
\midrule
\multicolumn{2}{l}{\textit{Horizon curriculum}}\\
Initial $\rightarrow$ max horizon & 4 $\rightarrow$ 20 quarters \\
Step & $+4$ per epoch after epoch 2 \\
Lookback & 12 quarters, fixed \\
\midrule
\multicolumn{2}{l}{\textit{Masking probabilities (per training example)}}\\
Mask entire future block & 0.5 \\
\quad else mask each future tuple & 0.95 \\
Select historical identity instance & 0.1 (mask $k\sim\mathrm{U}\{2,\dots,N\}$ of its $N$ members) \\
Mask ungrouped historical tuple & 0.05 \\
\bottomrule
\end{tabular}
\end{table}

\paragraph{Embeddings.} All embedding components share $d_{\mathrm{model}}=128$: a learned account-id embedding over the build's account vocabulary, a learned industry embedding over the 48 Fama--French industries plus an Unknown bucket, a fixed (parameter-free) sinusoidal encoding of the quarter offset relative to the origin, a learned value direction $w_x$ that scales with the standardized value, and a learned mask vector substituted for the value component of hidden tuples. The industry token is never masked and carries no prediction target.

\paragraph{Output heads.} The final representation of each queried tuple is concatenated with its sinusoidal horizon encoding (width $2d_{\mathrm{model}}=256$) and passed through two parallel two-layer MLPs ($256\rightarrow256\rightarrow1$, GELU): one for the mean $\mu$ and one for the log-variance $\log\sigma^2$. The log-variance is clamped to $[-10,10]$ before exponentiation; there is no other floor on $\sigma$.

\paragraph{Loss.} With $\beta=0.5$ the per-tuple Gaussian loss is
\[
\ell \;=\; \Big[\tfrac{(y-\mu)^2}{2\sigma^2} + \tfrac12\log\sigma^2\Big]
\cdot \big\lfloor \sigma^2 \big\rfloor_{\mathrm{sg}}^{\,\beta},
\]
where $y$ is the standardized target value of \S\ref{app:transformation}, $\lfloor\cdot\rfloor_{\mathrm{sg}}$ denotes stop-gradient \citep{seitzer2022pitfalls}, and $\beta=0$ recovers the standard NLL. The loss averages over tuples that are both masked and observed: synthetic query tokens without realized values (below) and the industry token never enter the loss. The absolute-error-track variant is a separately trained five-seed family, identical except that the head is read as a Laplace scale $b=\sigma/\sqrt2$ and trained with the Laplace $\beta$-NLL.

\paragraph{Masking procedure.} For each training example: (i) with probability 0.5 the entire future block is masked (pure forecasting); otherwise each future tuple is masked independently with probability 0.95, revealing $\approx$5\% of realized future values as inputs (the mechanism that lets a user pin assumed future values and condition the remaining forecasts on them; see the main paper's scenario analysis); (ii) each complete historical identity instance (all member accounts reported in a quarter) is selected with probability 0.1, and a selected instance masks $k$ members with $k$ drawn uniformly from $\{2,\dots,N\}$; $k$ is never exactly one, so no masked value is recoverable from its own identity; (iii) historical tuples of \emph{accounts} that participate in no complete historical identity instance in the window are masked independently with probability 0.05. An account appearing in a complete instance in any historical quarter is exempt from this singleton masking at all its quarters.

\paragraph{Curriculum.} Training runs 12 epochs with the maximum queried horizon following the schedule 4, 4, 8, 12, 16, 20,$\dots$, 20 (the first expansion comes after the second epoch); the training loader is rebuilt each epoch at the current horizon. The canonical run trains on the merged train and validation panel (1971--2009), like the refit tabular competitors of \S\ref{app:competitors}: all hyperparameters were fixed beforehand from validation-split experiments, no validation pass runs during training, and final-epoch weights are used. The 12-quarter lookback is fixed for every example; the 4-quarter minimum history only gates which firm-quarters are eligible origins.

\paragraph{Batching.} One example is the full tuple set of one firm-origin. No maximum sequence length is imposed and no example is truncated; sets are batched 32 at a time and padded to the longest set in the batch with a boolean key-padding attention mask. The average origin carries $\approx$758 observed historical tuples and $\approx$1{,}124 realized future targets; the model's token set additionally includes the synthetic query tokens for unrealized (account, horizon) cells (below), the scale token of \S\ref{app:transformation}, and one industry token.

\paragraph{Ex-ante future grid.} Future queries are built ex ante: for each origin, a synthetic query tuple is created for every (account, horizon) pair with the account drawn from the set the firm reports \emph{at the origin quarter} and the horizon running $1,\dots,20$, truncated only where the horizon extends past the split's last calendar quarter. That cut is identical for every firm and does not depend on the firm's own survival. Queries whose realization is later observed are loss-eligible targets; queries never realized (e.g.\ the firm exits) remain in the set as masked tokens but are excluded from the loss and masked at scoring time.

\paragraph{Inference.} At test time all future tuples are masked and a single forward pass yields $(\mu,\sigma)$ for every queried (item, horizon) pair jointly, with no iterated roll-forward. Forecasts are produced in fp32; $\sigma=\exp(\tfrac12\log\sigma^2)$ is saved alongside $\mu$. No reconciliation is applied at inference; the ex-post projections studied in the main paper are computed downstream from the saved forecasts.

\paragraph{Mixture.} The five seeds are trained independently and combined as an equal-weight mixture. Distributional metrics use the exact mixture: log-scores from the mixture density and CRPS from closed-form mixture expressions.

\paragraph{Compute.} Each seed trains in $\approx$16.5 hours on a single NVIDIA A100-SXM GPU; the five-seed run cost $\approx$\$120 of on-demand cloud compute. A trained checkpoint is $\approx$11\,MB.

\subsection{Competitor specifications}\label{app:competitors}

All learned tabular competitors share one protocol. They consume the tabular view: for each of the 78 items, four standardized recent levels and eight standardized year-over-year changes, plus the standardized deflator (\S\ref{app:transformation}) and 48 industry dummies, with the Unknown bucket as the omitted reference (985 features in all). Missing features are imputed with the cross-sectional latent-factor model of \citet{bryzgalova2025missing} with 10 factors, fit one calendar quarter at a time and filling only unobserved cells. Hyperparameters are selected on the 2002--2009 validation split (never test) and final models are refit on train$+$validation.

\paragraph{Elastic net.} One \texttt{sklearn} elastic net per (item, horizon) pair (1{,}560 fits; rows with any missing feature or target are dropped per fit). Grid: $\alpha\in\{0.01,0.1,1,10,100,1000\}\times \ell_1\text{-ratio}\in\{0.01,0.1,0.3,0.5,0.7,0.9,0.99\}$, selected by validation MSE per horizon on a reference item (net income) and shared across items at that horizon; coordinate descent, max 10{,}000 iterations.

\paragraph{Random forest.} One GPU (cuML) regression forest per (item, horizon) pair. Fixed parameters: 25 minimum samples per leaf, 50\% bootstrap per tree. Grid (same selection protocol as the elastic net): trees $\in\{50,100,200\}$, \texttt{max\_depth} $\in\{5,10,20\}$, \texttt{min\_samples\_split} $\in\{100,200\}$. The full 1{,}560-model fit runs on the order of 40 hours on one GPU.

\paragraph{Feed-forward networks.} Each FFNN is a single multi-output network mapping the 985 features to two 1{,}560-wide heads (a mean and a log-variance per (item, horizon)), trained with the same masked Gaussian $\beta$-NLL ($\beta=0.5$, $\log\sigma^2$ clamped to $[-10,10]$) as \textsc{Forma}; missing targets contribute zero loss rather than dropping the row, and rows are dropped only for residual missing features (rare after imputation). The \emph{linear} variant has no hidden layer (3{,}076{,}320 parameters); the \emph{large} variant has GELU hidden layers of widths 1024/1024/512 with dropout 0.1 (4{,}184{,}624 parameters), 3.3 and 4.4 times \textsc{Forma}'s 942{,}210 respectively. AdamW, learning rate $10^{-4}$, no weight decay, batch 1024; trained for a fixed 6 epochs (the best-validation epoch from a held-out selection run) on train$+$validation. Like \textsc{Forma}, each variant is a five-seed (60--64) equal-weight mixture with saved $\sigma$, so both enter the density track.

\paragraph{Chained GBM.} Our re-implementation of the chained gradient-boosting approach of \citet{geertsema2026chained} under the common protocol. LightGBM regressors are organized in a 14-step chain over 28 of the original specification's 29 core items (\texttt{iva} is dropped as scarcely populated in the quarterly file); the chain runs revenue first, then costs and working capital, through the capital-structure and tax blocks. At each step the feature matrix is augmented with the chain's upstream items: realized values during training (teacher forcing), predicted values at inference (cascade), with unreported upstream values entering as zero. Per item: up to 500 trees with early stopping (patience 10) on the validation split, leaves $\in\{15,31\}$ selected on validation, refit on train$+$validation at the selected size. Two arms are trained, an $L_1$ objective (the original specification, absolute-error track) and an $L_2$ objective (squared-error track). Of the 28 chained items, 25 lie in the 78-item \benchname{} universe; the headline table's Geert.\ column is scored on this 25-item footprint.

\paragraph{Pooled fade/AR(1).} For each (item, horizon), one pooled OLS regression of the standardized future value on the standardized origin value, $\hat x_{t+h} = \hat\alpha_{\mathrm{id},h} + \hat\rho_{\mathrm{id},h}\, x_t$, fit on train$+$validation (direct per-horizon regressions, never iterated), requiring at least 100 observations.

\paragraph{Seasonal random walk.} $\hat x_{t+h} = x_{t-((4-h)\bmod 4)}$: the forecast repeats the most recent observation falling in the same fiscal quarter, so the base sits $4$, $8$, $12$, $16$, or $20$ quarters before the target, and at $h\in\{4,8,12,16,20\}$ it coincides with the no-change forecast $\hat x_{t+h}=x_t$. Quarterly accounting series are seasonally differenced~\cite{foster1977quarterly}, and the seasonal random walk is the standard naive expectation against which the quarterly-earnings literature measures skill~\cite{bernard1990evidence}.

\paragraph{Chronos-2.} The public \texttt{amazon/chronos-2} checkpoint, zero-shot (no fine-tuning; it applies only its own inference-time normalization). For each origin the model receives the firm's raw (undeflated) reported values on a contiguous quarter grid, up to 12 observations through the origin with gaps entered as missing, which Chronos-2 handles natively, and forecasts all items jointly via its multivariate group attention. Its 21 native quantile levels $\{0.01, 0.05, 0.10, \dots, 0.90, 0.95, 0.99\}$ are mapped into benchmark space through the origin-frozen monotone transform of \S\ref{app:transformation}. The squared-error track uses the conditional mean obtained by trapezoidal integration over the mapped quantile grid (flat tails beyond the 0.01/0.99 levels); the absolute-error track uses the mapped median; the density track scores Chronos-2 directly from the same mapped quantile grid, CRPS by the exact piecewise-linear quantile expression and interval coverage from the matching quantile pairs. No log score is reported: on flat contexts ($\approx$0.6\% of cells) the predicted quantiles collapse to zero-width intervals that admit no usable density even under a scale floor.

\paragraph{LLM panel.} The frontier-LLM protocol and detailed prompts are documented in \apxref{app:llm}.

\clearpage

\section{LLM Benchmark}\label{app:llm}

The LLM panel is evaluated on the same information set as \textsc{Forma} and scored on the same grid, so differences across models reflect the forecaster rather than the information set. \S\ref{app:llm-protocol} documents the protocol; \S\ref{app:llm-prompts} reproduces the elicitation prompts.

\subsection{Protocol}\label{app:llm-protocol}

\paragraph{Inputs (tuple-sourced).} History and origins are read directly from \textsc{Forma}'s stored test tuples, with no re-derivation of any input \textsc{Forma} receives. Each prompt carries a 12-quarter reported history in original units, millions of dollars ($Q_{-11}\dots Q_0$, \textsc{Forma}'s \texttt{max\_lookback}), with Compustat mnemonic labels and \emph{relative} quarter indices, plus the firm's Fama--French-48 industry rendered as a plain-English sector name. The industry label gives the LLMs the same conditioning as \textsc{Forma}'s industry token and the baselines' dummies, and each of the 48 buckets is shared by hundreds of firms. The mnemonics are labels, not assumed knowledge: the system prompt carries a glossary giving every item's plain-English description and sign convention, plus the formulas of the derived subtotals shown as context (\S\ref{app:llm-prompts}). No firm identifier and no absolute dates enter the prompt. Returned forecasts are standardized with the origin-frozen parameters of \S\ref{app:transformation} before scoring.

\paragraph{Origins and sampler.} An eligible origin $(\text{firm},q_0)$ has $q_0$ in the test window 2010Q1--2024Q4, a valid scale at $q_0$, at least four quarters of history, and at least one future quarter to score; the eligible pool is 321{,}892 origins across 12{,}230 firms (the deflator-valid test firm-quarters of \S\ref{app:pipeline} that also pass these gates). For each sampled firm the sampler draws one variable-length block of \emph{consecutive} eligible origins (length 4--20), calendar-balanced across 2010--2024; consecutive origins produce the overlapping-target-quarter panels that a companion analysis of forecast revisions requires. One shared, seeded origins file is reused by every model to maximize pairwise overlap. The scored run is 2{,}103 pinned origins across 133 firms, run in three nested stages (the first 20, then 220, then all 2{,}103 origins of the shared file); the two prompt arms are paired on the identical origin set.

\paragraph{Targets.} The model forecasts the 55 primitive line items (only the firm's reported subset; structurally absent items are excluded from its list and treated as zero when reconstructing derived items). The remaining 23 of the 78 targets are reconstructed mechanically from the primitives through the accounting identities of \S\ref{app:identities}: statement subtotals such as \texttt{atq}, \texttt{ltq}, \texttt{seqq}, and \texttt{niq}. Eighteen of the 23 appear in the prompt's glossary as labeled derived context; the remaining five (\texttt{ancq}, \texttt{wcapq}, and the three carve-outs of \S\ref{app:pipeline}) are reconstructed downstream and never shown to the model. Every model is therefore scored on the same 78-target~$\times$~$Q_1$--$Q_{20}$ grid as \textsc{Forma}. Every origin is asked for the full $Q_1$--$Q_{20}$ horizon regardless of how many future quarters the firm actually has in file; unavailable horizons are masked only at scoring time, so the prompt never leaks the firm's data-end or delisting.

\paragraph{Two elicitation arms.} Both arms share the same inputs, the same 55-primitive schema, and the same strict primitives-only output format; they differ only in one system-prompt block. The \emph{unstructured} arm pairs the line-item glossary and accounting-consistency checks with a free-form steering paragraph that tells the model to reason about trajectory, growth, seasonality, mean-reversion, the identities, and the drivers, without prescribing a method. The \emph{structured} arm replaces that paragraph with an explicit driver-hierarchy / roll-forward method (revenue year-over-year off the same quarter a year earlier, cost and expense lines as ratios to revenue, working capital via turnover ratios, PP\&E by roll-forward, then cash and retained earnings \emph{set} by the cash-flow and retained-earnings rolls), using the identities as the generative construction order. The comparison therefore carries no parse-format risk. The headline table reports each model's better arm, the unstructured arm for all three models. Unstructured/structured change-space $R^2$ is 0.186/0.175 for Opus~4.8, 0.174/0.171 for GPT-5.5, and 0.158/0.139 for Sonnet~5; MAE is 0.362/0.367, 0.363/0.368, and 0.368/0.376.

\paragraph{Models and settings.} The panel comprises three frontier models, Claude Opus~4.8, GPT-5.5, and Claude Sonnet~5, all with extended reasoning enabled. The Anthropic models run at their adaptive default (bounded, measured at ${\approx}4$--$6$k thinking tokens for Opus and ${\approx}12$--$17$k for Sonnet~5); GPT-5.5 runs at medium reasoning effort. Reasoning is therefore not token-matched across vendors (the batch gateway exposes no common reasoning-effort setting), a run property to bear in mind when comparing models. All calls use default temperature and \texttt{max\_tokens}${=}64{,}000$ (headroom for heavy reasoning on the structured prompt; billed only when consumed) and route through the Portkey batch gateway: the Anthropic models via the Messages batch endpoint, and GPT-5.5's scored run via the OpenAI Responses endpoint to capture reasoning summaries ($\approx$82\% of origins carry one). The main run is a single call per origin.

\paragraph{Decoding-noise subset.} A 200-origin simple-random subset (drawn from the main sample) is queried $K{=}5$ times on every model and both arms at default temperature to measure across-call dispersion; the main run's single pass serves as draw~1, so the extra cost is four replicate runs per cell. The subset supports the companion analysis of forecast revisions and does not enter the headline results.

\paragraph{Common sample.} The LLM column of the headline table is an exact common sample: a (firm, origin, item, horizon) cell counts only where all seven forecasters (\textsc{Forma} and the six LLM runs, two prompt arms $\times$ three models) predict it finitely and a realized value is available to score against. That intersection is 2{,}152{,}314 cells, of a possible $78 \times 20 \times 2{,}103 = 3{,}280{,}680$ (targets $\times$ horizons $\times$ origins), a 66\% subset; the full model suite is then re-scored inside that footprint. Parse failures are essentially absent (every model returned forecasts on $\approx$100\% of requested cells), so exclusions come from cells with no realized value to score or outside a model's covered item set at that origin.

\subsection{Exact prompts}\label{app:llm-prompts}

Prompts are reproduced from the released prompt files with Unicode box-drawing and math glyphs transliterated to ASCII for typesetting; the byte-exact originals are included in the release. The \texttt{\{lookback\_qs\}} / \texttt{\{lookback\_first\}} / \texttt{\{max\_horizon\}} tokens are substituted from config at load time and render as \texttt{12} / \texttt{11} / \texttt{20} for the production run.

\subsubsection{System prompt, unstructured arm, in full}
\begin{Verbatim}
You are a financial forecasting model. Given {lookback_qs} quarters of historical
financial data (Q-{lookback_first} through Q0) for one company, forecast forward
quarters Q1 through Q{max_horizon}.

Forecasting these line items accurately is a hard, multi-step reasoning
problem. Before committing to any numbers, reason carefully and thoroughly:
study the firm's recent trajectory and growth rates, seasonality, mean-
reversion, the accounting identities that link the statements, and the likely
path of each driver. Think deeply and check your work before answering. Do all
of this reasoning internally -- the visible output must still be ONLY the
Q-lines specified in the OUTPUT FORMAT section below.

Each user message states the firm's industry using the Fama-French 48 sector
classification. Treat it as a prior -- sector-typical margins, seasonality,
asset intensity, capital structure, and growth differ markedly across
industries -- but always defer to the firm's own reported history where the two
disagree.

All values are in millions USD. Negative = losses / outflows / contra-equity.
Variable names are Compustat quarterly codes. All cash-flow items are
QUARTERLY (pipeline has already converted YTD reporting to single-quarter
flows). Signs follow Compustat conventions -- SEE THE SIGN COLUMN in each
glossary row.

=============================================================================
GLOSSARY -- primitives you MUST forecast are marked [P]; derived items are
computed mechanically from primitives and are shown here as context only.
=============================================================================

-- INCOME STATEMENT ---------------------------------------------------------
  [P] revtq    Revenue, total                                           (+)
  [P] cogsq    Cost of goods sold                                       (+)
  [P] xsgaq    Selling, general & admin expense -- INCLUDES xrdq         (+)
  [P] xrdq     Research & development expense (informational sub of xsgaq) (+)
  [P] dpq      Depreciation & amortization expense                      (+)
  [P] stkcoq   Stock-based compensation expense (informational)         (+)
  [P] xintq    Interest expense                                         (+)
  [P] nopiq    Non-operating income/(expense), net                    (signed)
  [P] spiq     Special items (impairments, restructuring, etc)        (signed; usually <= 0)
  [P] txtq     Income tax expense                                     (signed)
  [P] miiq     Noncontrolling interest (income attrib. to minority)   (signed)
  [P] xidoq    Extraordinary items & discontinued operations          (signed)
      gpq      Gross profit           = revtq - cogsq                  (derived)
      oibdpq   EBITDA                 = gpq - xsgaq                    (derived)
      oiadpq   EBIT                   = oibdpq - dpq                   (derived)
      piq      Pretax income          = oiadpq - xintq + nopiq + spiq  (derived)
      ibq      Income before extra    = piq - txtq - miiq              (derived)
      niq      Net income             = ibq + xidoq                    (derived)
      xoprq    Total operating exp.   = cogsq + xsgaq                  (derived)

-- BALANCE SHEET -- ASSETS ---------------------------------------------------
  [P] cheq     Cash & short-term investments                            (+)
  [P] rectq    Accounts receivable, trade                               (+)
  [P] invtq    Inventories                                              (+)
  [P] acoq     Other current assets                                     (+)
  [P] ppentq   Property, plant & equipment, NET                         (+)
  [P] dpactq   Accumulated depreciation                                 (+)
  [P] gdwlq    Goodwill                                                 (+)
  [P] intanoq  Other intangibles (excl. goodwill)                       (+)
  [P] aoq      Other assets -- AGGREGATE that already contains intanq    (+)
      intanq   Intangibles total      = gdwlq + intanoq                 (derived)
      ppegtq   PP&E gross             = ppentq + dpactq                 (derived)
      actq     Current assets total   = cheq + rectq + invtq + acoq     (derived)
      atq      Total assets           = actq + ppentq + aoq             (derived)

-- BALANCE SHEET -- LIABILITIES ----------------------------------------------
  [P] apq      Accounts payable, trade                                  (+)
  [P] dlcq     Debt in current liabilities (short-term debt)            (+)
  [P] txpq     Income taxes payable                                     (+)
  [P] drcq     Deferred revenue, current                                (+)
  [P] lcoq     Other current liabilities                                (+)
  [P] drltq    Deferred revenue, long-term -- subset of loq              (+)
  [P] dlttq    Long-term debt                                           (+)
  [P] txditcq  Deferred taxes & investment tax credit                 (signed; usually +)
  [P] loq      Other liabilities -- AGGREGATE that already contains drltq (+)
      lctq     Current liab. total    = apq + dlcq + txpq + lcoq        (derived)
                                        (drcq is typically inside lcoq;
                                         do not add drcq again)
      ltq      Total liabilities      = lctq + dlttq + txditcq + loq    (derived)

-- BALANCE SHEET -- EQUITY ---------------------------------------------------
  [P] cstkq    Common stock, par value                                  (+)
  [P] capsq    Additional paid-in capital / capital surplus             (+)
  [P] req      Retained earnings -- already includes AOCI              (signed)
  [P] acomincq AOCI component inside req (disclosure breakout)        (signed)
  [P] tstkq    Treasury stock at cost -- POSITIVE magnitude, subtracted  (+)
  [P] pstkq    Preferred stock                                          (+)
  [P] mibtq    Noncontrolling interest, total                           (+)
      ceqq     Common equity          = cstkq + capsq + req - tstkq     (derived)
      seqq     Total stockholders eq. = ceqq + pstkq                    (derived)

-- CASH FLOW (all quarterly; YTD already converted) -------------------------
Operating:
  [P] oancfq   Cash from operating activities                         (signed)
  [P] fopoq    Other operating CF adjustments                         (signed)
Investing (individual components; ivncfq is DERIVED from them):
  [P] capxq    Capital expenditures -- positive OUTFLOW                  (+)
  [P] ivchq    Increase in LT investments -- positive OUTFLOW            (+)
  [P] aqcq     Acquisitions -- positive OUTFLOW                          (+)
  [P] sivq     Sale of investments -- positive INFLOW                    (+)
  [P] sppeq    Sale of property -- positive INFLOW                       (+)
  [P] ivstchq  Change in ST investments (separate from ivchq)         (signed)
  [P] ivacoq   Other investing activities                             (signed)
      ivncfq   Cash from investing    = -capxq - ivchq - aqcq
                                        + sivq + sppeq + ivstchq + ivacoq (derived)
Financing (individual components; fincfq is DERIVED from them):
  [P] sstkq    Stock issuance -- positive INFLOW                         (+)
  [P] prstkcq  Stock repurchases -- positive OUTFLOW                     (+)
  [P] dltisq   Long-term debt issuance -- positive INFLOW                (+)
  [P] dltrq    Long-term debt repayment -- positive OUTFLOW              (+)
  [P] dlcchq   Changes in short-term debt                             (signed)
  [P] dvq      Dividends paid -- positive OUTFLOW                        (+)
  [P] txbcofq  Excess tax benefit of stock options                    (signed; often 0 post-2016)
  [P] fiaoq    Other financing activities                             (signed)
      fincfq   Cash from financing    = sstkq - prstkcq + dltisq - dltrq
                                        + dlcchq - dvq + txbcofq + fiaoq (derived)
FX & free cash flow:
  [P] exreq    Effect of exchange rate on cash                        (signed)
      fcfq     Free cash flow         = oancfq - capxq                  (derived)

=============================================================================
ACCOUNTING CONSISTENCY CHECKS (your primitives must satisfy these)
=============================================================================

  (1) Accounting equation:   atq = ltq + mibtq + seqq
  (2) Cash rollforward:      cheq[t] = cheq[t-1] + oancfq + ivncfq + fincfq + exreq
  (3) Retained earnings:     req[t] ~= req[t-1] + niq - dvq + Delta acomincq
                             (approximate; non-cash AOCI + reclassifications add noise)

=============================================================================
OUTPUT FORMAT
=============================================================================

Each user message lists the EXACT subset of [P] items this firm reports
(items it never discloses are excluded -- they are treated as 0 by us). You
must:
  * Forecast EVERY item in that per-firm list.
  * Do NOT output any item not in that list (no hallucinated zeros for
    structurally-absent items).
  * Do NOT output any derived item -- they are recomputed mechanically.

One line per horizon, values comma-separated. Output Q1 through Q{max_horizon}
(forecast EVERY horizon in that range, even if you are uncertain a later
quarter will materialize -- do not stop early):

Q1: revtq=X.XX, cogsq=X.XX, xsgaq=X.XX, ...
Q2: revtq=X.XX, cogsq=X.XX, xsgaq=X.XX, ...
...
Q{max_horizon}: revtq=X.XX, cogsq=X.XX, xsgaq=X.XX, ...

Output ONLY the Q-line forecasts. No markdown, no commentary, no code fences.
\end{Verbatim}

\subsubsection{Structured arm: the block that replaces the steering paragraph}
\noindent\textit{The remainder of the prompt (industry prior, units and sign conventions, glossary, consistency checks, output format) is unchanged from the unstructured arm.}
\begin{Verbatim}
You are a financial forecasting model. Given {lookback_qs} quarters of historical
financial data (Q-{lookback_first} through Q0) for one company, forecast forward
quarters Q1 through Q{max_horizon}.

PROJECTION METHOD -- forecast a few DRIVERS, then DERIVE the rest. Reason
through this sequence internally (in growth rates and ratios), then convert
back to dollar levels before emitting:

  1. Revenue (revtq) first. Anchor on the SAME fiscal quarter a year earlier:
     revtq[t] ~= revtq[t-4] x (1 + g), where g is a year-over-year growth rate
     read from the recent trailing-4-quarter trend and FADED toward a modest
     long-run rate across the 20-quarter horizon -- do not let a recent spike
     persist to Q20. Anchoring on t-4 preserves seasonality.

  2. Operating lines as ratios to revenue. Project cogsq, xsgaq (which
     includes xrdq), dpq, and stkcoq each as a share of revtq, held near its
     trailing-4-quarter median and drifting only with a clear operating-
     leverage reason; convert back to a level = ratio x revtq[t]. Income tax
     (txtq) ~= tax rate x pretax income, and tracks the sign of pretax income.

  3. Working capital via turnover. Project receivables (rectq) off revtq
     (days-sales-outstanding) and inventory (invtq) and payables (apq) off
     cogsq (days-inventory / days-payable); hold each turnover near its
     trailing level, then convert back to a level using projected revtq/cogsq.

  4. PP&E by roll-forward. ppentq[t] ~= ppentq[t-1] + capxq - dpq, with capxq
     projected as a share of revtq.

  5. Cash flow first, then SET cash and retained earnings by roll-forward --
     do NOT forecast cheq or req independently:
       * Operating cash flow must be consistent with earnings and working
         capital: oancfq ~= niq + dpq - Delta rectq - Delta invtq + Delta apq. A RISE in
         receivables or inventory is a USE of cash (subtract); a rise in
         payables is a SOURCE (add); dpq is a positive add-back. capxq is a
         cash OUTFLOW (in ivncfq); debt issuance is a source while repayments
         and dividends (dvq) are uses (in fincfq).
       * Cash: cheq[t] = cheq[t-1] + oancfq + ivncfq + fincfq + exreq. Set
         cheq to this rollforward result, not to an independent guess.
       * Retained earnings: req[t] ~= req[t-1] + niq - dvq + Delta acomincq. Set req
         to this roll.

Before emitting each quarter, silently confirm the three ACCOUNTING
CONSISTENCY identities below hold (accounting equation, cash rollforward,
retained-earnings roll); if one fails, adjust the dependent line (cheq, req,
or an equity line) so it holds.

Do all of this reasoning internally -- the visible output must still be ONLY
the Q-lines specified in the OUTPUT FORMAT section below.

Each user message states the firm's industry using the Fama-French 48 sector
classification. Treat it as a prior -- sector-typical margins, seasonality,
asset intensity, capital structure, and growth differ markedly across
industries -- but always defer to the firm's own reported history where the two
disagree.

All values are in millions USD. Negative = losses / outflows / contra-equity.
Variable names are Compustat quarterly codes. All cash-flow items are
QUARTERLY (pipeline has already converted YTD reporting to single-quarter
flows). Signs follow Compustat conventions -- SEE THE SIGN COLUMN in each
glossary row.

=============================================================================
GLOSSARY -- primitives you MUST forecast are marked [P]; derived items are
computed mechanically from primitives and are shown here as context only.
=============================================================================

-- INCOME STATEMENT ---------------------------------------------------------
  [P] revtq    Revenue, total                                           (+)
  [P] cogsq    Cost of goods sold                                       (+)
  [P] xsgaq    Selling, general & admin expense -- INCLUDES xrdq         (+)
  [P] xrdq     Research & development expense (informational sub of xsgaq) (+)
  [P] dpq      Depreciation & amortization expense                      (+)
  [P] stkcoq   Stock-based compensation expense (informational)         (+)
  [P] xintq    Interest expense                                         (+)
  [P] nopiq    Non-operating income/(expense), net                    (signed)
  [P] spiq     Special items (impairments, restructuring, etc)        (signed; usually <= 0)
  [P] txtq     Income tax expense                                     (signed)
  [P] miiq     Noncontrolling interest (income attrib. to minority)   (signed)
  [P] xidoq    Extraordinary items & discontinued operations          (signed)
      gpq      Gross profit           = revtq - cogsq                  (derived)
      oibdpq   EBITDA                 = gpq - xsgaq                    (derived)
      oiadpq   EBIT                   = oibdpq - dpq                   (derived)
      piq      Pretax income          = oiadpq - xintq + nopiq + spiq  (derived)
      ibq      Income before extra    = piq - txtq - miiq              (derived)
      niq      Net income             = ibq + xidoq                    (derived)
      xoprq    Total operating exp.   = cogsq + xsgaq                  (derived)

-- BALANCE SHEET -- ASSETS ---------------------------------------------------
  [P] cheq     Cash & short-term investments                            (+)
  [P] rectq    Accounts receivable, trade                               (+)
  [P] invtq    Inventories                                              (+)
  [P] acoq     Other current assets                                     (+)
  [P] ppentq   Property, plant & equipment, NET                         (+)
  [P] dpactq   Accumulated depreciation                                 (+)
  [P] gdwlq    Goodwill                                                 (+)
  [P] intanoq  Other intangibles (excl. goodwill)                       (+)
  [P] aoq      Other assets -- AGGREGATE that already contains intanq    (+)
      intanq   Intangibles total      = gdwlq + intanoq                 (derived)
      ppegtq   PP&E gross             = ppentq + dpactq                 (derived)
      actq     Current assets total   = cheq + rectq + invtq + acoq     (derived)
      atq      Total assets           = actq + ppentq + aoq             (derived)

-- BALANCE SHEET -- LIABILITIES ----------------------------------------------
  [P] apq      Accounts payable, trade                                  (+)
  [P] dlcq     Debt in current liabilities (short-term debt)            (+)
  [P] txpq     Income taxes payable                                     (+)
  [P] drcq     Deferred revenue, current                                (+)
  [P] lcoq     Other current liabilities                                (+)
  [P] drltq    Deferred revenue, long-term -- subset of loq              (+)
  [P] dlttq    Long-term debt                                           (+)
  [P] txditcq  Deferred taxes & investment tax credit                 (signed; usually +)
  [P] loq      Other liabilities -- AGGREGATE that already contains drltq (+)
      lctq     Current liab. total    = apq + dlcq + txpq + lcoq        (derived)
                                        (drcq is typically inside lcoq;
                                         do not add drcq again)
      ltq      Total liabilities      = lctq + dlttq + txditcq + loq    (derived)

-- BALANCE SHEET -- EQUITY ---------------------------------------------------
  [P] cstkq    Common stock, par value                                  (+)
  [P] capsq    Additional paid-in capital / capital surplus             (+)
  [P] req      Retained earnings -- already includes AOCI              (signed)
  [P] acomincq AOCI component inside req (disclosure breakout)        (signed)
  [P] tstkq    Treasury stock at cost -- POSITIVE magnitude, subtracted  (+)
  [P] pstkq    Preferred stock                                          (+)
  [P] mibtq    Noncontrolling interest, total                           (+)
      ceqq     Common equity          = cstkq + capsq + req - tstkq     (derived)
      seqq     Total stockholders eq. = ceqq + pstkq                    (derived)

-- CASH FLOW (all quarterly; YTD already converted) -------------------------
Operating:
  [P] oancfq   Cash from operating activities                         (signed)
  [P] fopoq    Other operating CF adjustments                         (signed)
Investing (individual components; ivncfq is DERIVED from them):
  [P] capxq    Capital expenditures -- positive OUTFLOW                  (+)
  [P] ivchq    Increase in LT investments -- positive OUTFLOW            (+)
  [P] aqcq     Acquisitions -- positive OUTFLOW                          (+)
  [P] sivq     Sale of investments -- positive INFLOW                    (+)
  [P] sppeq    Sale of property -- positive INFLOW                       (+)
  [P] ivstchq  Change in ST investments (separate from ivchq)         (signed)
  [P] ivacoq   Other investing activities                             (signed)
      ivncfq   Cash from investing    = -capxq - ivchq - aqcq
                                        + sivq + sppeq + ivstchq + ivacoq (derived)
Financing (individual components; fincfq is DERIVED from them):
  [P] sstkq    Stock issuance -- positive INFLOW                         (+)
  [P] prstkcq  Stock repurchases -- positive OUTFLOW                     (+)
  [P] dltisq   Long-term debt issuance -- positive INFLOW                (+)
  [P] dltrq    Long-term debt repayment -- positive OUTFLOW              (+)
  [P] dlcchq   Changes in short-term debt                             (signed)
  [P] dvq      Dividends paid -- positive OUTFLOW                        (+)
  [P] txbcofq  Excess tax benefit of stock options                    (signed; often 0 post-2016)
  [P] fiaoq    Other financing activities                             (signed)
      fincfq   Cash from financing    = sstkq - prstkcq + dltisq - dltrq
                                        + dlcchq - dvq + txbcofq + fiaoq (derived)
FX & free cash flow:
  [P] exreq    Effect of exchange rate on cash                        (signed)
      fcfq     Free cash flow         = oancfq - capxq                  (derived)

=============================================================================
ACCOUNTING CONSISTENCY CHECKS (your primitives must satisfy these)
=============================================================================

  (1) Accounting equation:   atq = ltq + mibtq + seqq
  (2) Cash rollforward:      cheq[t] = cheq[t-1] + oancfq + ivncfq + fincfq + exreq
  (3) Retained earnings:     req[t] ~= req[t-1] + niq - dvq + Delta acomincq
                             (approximate; non-cash AOCI + reclassifications add noise)

=============================================================================
OUTPUT FORMAT
=============================================================================

Each user message lists the EXACT subset of [P] items this firm reports
(items it never discloses are excluded -- they are treated as 0 by us). You
must:
  * Forecast EVERY item in that per-firm list.
  * Do NOT output any item not in that list (no hallucinated zeros for
    structurally-absent items).
  * Do NOT output any derived item -- they are recomputed mechanically.

One line per horizon, values comma-separated. Output Q1 through Q{max_horizon}
(forecast EVERY horizon in that range, even if you are uncertain a later
quarter will materialize -- do not stop early):

Q1: revtq=X.XX, cogsq=X.XX, xsgaq=X.XX, ...
Q2: revtq=X.XX, cogsq=X.XX, xsgaq=X.XX, ...
...
Q{max_horizon}: revtq=X.XX, cogsq=X.XX, xsgaq=X.XX, ...

Output ONLY the Q-line forecasts. No markdown, no commentary, no code fences.
\end{Verbatim}

\subsubsection{Per-origin user message (both arms)}
Assembled per origin: the fixed horizon instruction, the firm's Fama--French-48 industry, the firm's active-primitive list, then the 12-quarter history as CSV-like blocks (one per statement; rows null across all history are suppressed). Schematic, with illustrative values and the history truncated to 3 of 12 quarters for space:

\begin{Verbatim}[fontsize=\small,frame=single]
Forecast horizons: Q1 through Q20.

Firm industry (Fama-French 48): Retail.

This firm reports 47 primitive line items. Forecast ONLY these
items -- do not output any other primitives:

revtq, cogsq, xsgaq, dpq, xintq, ..., oancfq, capxq, dvq

Items not in this list are structurally absent for this firm
(consistently null in history); they will be treated as 0.

Historical data:

=== INCOME STATEMENT ===
Account,Q-11,...,Q-1,Q0
revtq,1203.40,...,1456.10,1502.77
cogsq,742.10,...,889.30,910.55
...
=== BALANCE SHEET - ASSETS ===
Account,Q-11,...,Q-1,Q0
cheq,310.22,...,402.10,419.88
...
\end{Verbatim}

\noindent The model returns one comma-separated line per horizon (\texttt{Q1: revtq=..., cogsq=..., \dots}\,$\rightarrow$\,\texttt{Q20: \dots}), primitives only; the 23 derived and subtotal items are reconstructed mechanically downstream.

\end{document}